# Multi-Instance Learning with Any Hypothesis Class

**Sivan Sabato** SIVAN_SABATO@CS.HUJI.AC.IL

**Naftali Tishby** TISHBY@CS.HUJI.AC.IL

*School of Computer Science & Engineering*
*The Hebrew University*
*Jerusalem 91904, Israel*

## Abstract

In the supervised learning setting termed Multiple-Instance Learning (MIL), the examples are bags of instances, and the bag label is a function of the labels of its instances. Typically, this function is the Boolean OR. The learner observes a sample of bags and the bag labels, but not the instance labels that determine the bag labels. The learner is then required to emit a classification rule for bags based on the sample. MIL has numerous applications, and many heuristic algorithms have been used successfully on this problem, each adapted to specific settings or applications. In this work we provide a unified theoretical analysis for MIL, which holds for any underlying hypothesis class, regardless of a specific application or problem domain. We show that the sample complexity of MIL is only poly-logarithmically dependent on the size of the bag, for any underlying hypothesis class. In addition, we introduce a new PAC-learning algorithm for MIL, which uses a regular supervised learning algorithm as an oracle. We prove that efficient PAC-learning for MIL can be generated from any efficient non-MIL supervised learning algorithm that handles one-sided error. The computational complexity of the resulting algorithm is only polynomially dependent on the bag size.

## 1. Introduction

We consider the learning problem termed Multiple-Instance Learning (MIL), first introduced in Dietterich et al. (1997). MIL is a special type of a supervised classification problem. As in classical supervised classification, in MIL the learner receives a sample of labeled examples drawn i.i.d from an arbitrary and unknown distribution, and its objective is to discover a classification rule with a small expected error over the same distribution. In MIL additional structure is assumed, whereby the examples are received as *bags* of *instances*, such that each bag is composed of several instances. It is assumed that each instance has a true label, however the learner only observes the labels of the bags. In classical MIL the label of a bag is the Boolean OR of the labels of the instances the bag contains. Various generalizations to MIL have been proposed (see e.g. Raedt, 1998; Weidmann et al., 2003). Here we consider both classical MIL and the more general setting, where a function other than Boolean OR determines bag labels based on instance labels. This function is known to the learner a-priori. We term the more general setting *generalized MIL*.

It is possible, in principle, to view MIL as a regular supervised classification task, where a bag is a single example, and the instances in a bag are merely part of its internal representation. Such a view, however, means that one must analyze each specific MIL problem separately, and that results and methods that apply to one MIL problem are not transferable to other MIL problems. We propose instead a generic approach to the analysis of MIL, in which the properties of a MIL problem are analyzed as a function of the properties of the matching non-MIL problem. As we show in this work, the connections between the MIL and the non-MIL properties are strong and useful. The generic approach has the advantage that it automatically extends all knowledge and methods that apply to non-MIL problems into knowledge and methods that apply to MIL, without requiring specialized analysis for each specific MIL problem. Our results are thus applicable for diverse hypothesis classes, label relationships between bags and instances, and target losses. Moreover, the generic approach allows a better theoretical understanding of the relationship, in general, between regular learning and multi-instance learning with the same hypothesis class.

The generic approach can also be helpful for the design of algorithms, since it allows deriving generic methods and approaches that hold across different settings. For instance, as we show below, a generic PAC-learning algorithm can be derived for a large class of MIL problems with different hypothesis classes. Other applications can be found in follow-up research of the results we report here, such as a generic bag-construction mechanism (Sabato et al., 2010), and learning when bags have a manifold structure (Babenko et al., 2011). As generic analysis goes, it might be possible to improve upon it in some specific cases. Identifying these cases and providing tighter analysis for them is an important topic for future work. We do show that in some important cases—most notably that of learning separating hyperplanes with classical MIL—our analysis is tight up to constants.

MIL has been used in numerous applications. In Dietterich et al. (1997) the drug design application motivates this setting. In this application, the goal is to predict which molecules would bind to a specific binding site. Each molecule has several possible conformations (shapes) it can take. If at least one of the conformations binds to the binding site, then the molecule is labeled positive. However, it is not possible to experimentally identify which conformation was the successful one. Thus, a molecule can be thought of as a bag of conformations, where each conformation is an instance in the bag representing the molecule. This application employs the hypothesis class of Axis Parallel Rectangles (APRs), and has made APRs the hypothesis class of choice in several theoretical works that we mention below. There are many other applications for MIL, including image classification (Maron and Ratan, 1998), web index page recommendation (Zhou et al., 2005) and text categorization (Andrews, 2007).

Previous theoretical analysis of the computational aspects of MIL has been done in two main settings. In the first setting, analyzed for instance in Auer et al. (1998); Blum and Kalai (1998); Long and Tan (1998), it is assumed that all the instances are drawn i.i.d from a single distribution over instances, so that the instances in each bag are statistically independent. Under this independence assumption, learning from an i.i.d. sample of bags is as easy as learning from an i.i.d. sample of instances with one-sided label noise. This is stated in the following theorem.

**Theorem 1 (Blum and Kalai, 1998)** *If a hypothesis class is PAC-learnable in polynomial time from one-sided random classification noise, then the same hypothesis class is PAC-learnable in polynomial time in MIL under the independence assumption. The computational complexity of learning is polynomial in the bag size and in the sample size.*

The assumption of statistical independence of the instances in each bag is, however, very limiting, as it is irrelevant to many applications.

In the second setting one assumes that bags are drawn from an arbitrary distribution *over bags*, so that the instances within a bag may be statistically dependent. This is clearly much more useful in practice, since bags usually describe a complex object with internal structure, thus it is implausible to assume even approximate independence of instances in a bag. For the hypothesis class of APRs and an arbitrary distribution over bags, it is shown in Auer et al. (1998) that if there exists a PAC-learning algorithm for MIL with APRs, and this algorithm is polynomial in both the size of the bag and the dimension of the Euclidean space, then it is possible to polynomially PAC-learn DNF formulas, a problem which is solvable only if $\mathcal{RP} = \mathcal{NP}$ (Pitt and Valiant, 1986). In addition, if it is possible to improperly learn MIL with APRs (that is, to learn a classifier which is not itself an APR), then it is possible to improperly learn DNF formulas, a problem which has not been solved to this date for general distributions. This result implies that it is not possible to PAC-learn MIL on APRs using an algorithm which is efficient in both the bag size and the problem's dimensionality. It does not, however, preclude the possibility of performing MIL efficiently in other cases.

In practice, numerous algorithms have been proposed for MIL, each focusing on a different specialization of this problem. Almost none of these algorithms assume statistical independence of instances in a bag. Moreover, some of the algorithms explicitly exploit presumed dependences between instances in a bag. Dietterich et al. (1997) propose several heuristic algorithms for finding an APR that predicts the label of an instance and of a bag. Diverse Density (Maron and Lozano-Pérez, 1998) and EM-DD (Zhang and Goldman, 2001) employ assumptions on the structure of the bags of instances. DPBoost (Andrews and Hofmann, 2003), mi-SVM and MI-SVM (Andrews et al., 2002), and Multi-Instance Kernels (Gärtner et al., 2002) are approaches for learning MIL using margin-based objectives. Some of these methods work quite well in practice. However, no generalization guarantees have been provided for any of them.

In this work we analyze MIL and generalized MIL in a general framework, independent of a specific application, and provide results that hold for any underlying hypothesis class. We assume a fixed hypothesis class defined over instances. We then investigate the relationship between learning with respect to this hypothesis class in the classical supervised learning setting with no bags, and learning with respect to the same hypothesis class in MIL. We address both sample complexity and computational feasibility.

Our sample complexity analysis shows that for binary hypothesis and thresholded real-valued hypotheses, the distribution-free sample complexity for generalized MIL grows only logarithmically with the maximal bag size. We also provide poly-logarithmic sample complexity bounds for the case of margin learning. We further provide distribution-dependent sample complexity bounds for more general loss functions. These bound are useful when only the average bag size is bounded. The results imply generalization bounds for previously proposed algorithms for MIL. Addressing the computational feasibility of MIL, we provide a new learning algorithm with provable guarantees for a class of bag-labeling functions that includes the Boolean OR, used in classical MIL, as a special case. Given a non-MIL learning algorithm for the desired hypothesis class, which can handle one-sided errors, we improperly learn MIL with the same hypothesis class. The construction is simple to implement, and provides a computationally efficient PAC-learning of MIL, with only a polynomial dependence of the run time on the bag size.

In this work we consider the problem of learning to classify bags using a labeled sample of bags. We do not attempt to learn to classify single instances using a labeled sample of bags. We point out that it is not generally possible to find a low-error classification rule for instances based on a bag sample. As a simple counter example, assume that the label of a bag is the Boolean OR of the labels of its instances, and that every bag includes both a positive instance and a negative instance. In this case all bags are labeled as positive, and it is not possible to distinguish the two types of instances by observing only bag labels.

The structure of the paper is as follows. In Section 2 the problem is formally defined and notation is introduced. In Section 3 the sample complexity of generalized MIL for binary hypotheses is analyzed. We provide a useful lemma bounding covering numbers for MIL in Section 4. In Section 5 we analyze the sample complexity of generalized MIL with real-valued functions for large-margin learning. Distribution-dependent results for binary learning and real-valued learning based on the average bag size are presented in Section 6. In Section 7 we present a PAC-learner for MIL and analyze its properties. We conclude in Section 8. The appendix includes technical proofs that have been omitted from the text. A preliminary version of this work has been published as Sabato and Tishby (2009).

## 2. Notations and Definitions

For a natural number $k$, we denote $[k] \triangleq \{1, \ldots, k\}$. For a real number $x$, we denote $[x]_+ = \max\{0, x\}$. $\log$ denotes a base 2 logarithm. For two vectors $\mathbf{x}, \mathbf{y} \in \mathbb{R}^n$, $\langle \mathbf{x}, \mathbf{y} \rangle$ denotes the inner product of $\mathbf{x}$ and $\mathbf{y}$. We use the function $\mathrm{sign} : \mathbb{R} \to \{-1, +1\}$ where $\mathrm{sign}(x) = 1$ if $x \geq 0$ and $\mathrm{sign}(x) = -1$ otherwise. For a function $f : A \to B$, we denote by $f_{|C}$ its restriction to a set $C \subseteq A$. For a univariate function $f$, denote its first and second derivatives by $f'$ and $f''$ respectively.

Let $\mathcal{X}$ be the input space, also called the domain of instances. A bag is a finite ordered set of instances from $\mathcal{X}$. Denote the set of allowed sizes for bags in a specific MIL problem by $R \subseteq \mathbb{N}$. For any set $A$ we denote $A^{(R)} \triangleq \cup_{n \in R} A^n$. Thus the domain of bags with a size in $R$ and instances from $\mathcal{X}$ is $\mathcal{X}^{(R)}$. A bag of size $n$ is denoted by $\bar{\mathbf{x}} = (x[1], \ldots, x[n])$ where each $x[j] \in \mathcal{X}$ is an instance in the bag. We denote the number of instances in $\bar{\mathbf{x}}$ by $|\bar{\mathbf{x}}|$. For any univariate function $f : A \to B$, we may also use its extension to a multivariate function from sequences of elements in $A$ to sequences of elements in $B$, defined by $f(a[1], \ldots, a[k]) = (f(a[1]), \ldots, f(a[k]))$.

Let $I \subseteq \mathbb{R}$ an allowed range for hypotheses over instances or bags. For instance, $I = \{-1, +1\}$ for binary hypotheses and $I = [-B, B]$ for real-valued hypotheses with a bounded range. $\mathcal{H} \subseteq I^{\mathcal{X}}$ is a hypothesis class for instances. Every MIL problem is defined by a fixed bag-labeling function $\psi : I^{(R)} \to I$ that determines the bag labels given the instance labels. Formally, every instance hypothesis $h : \mathcal{X} \to I$ defines a bag hypothesis, denoted by $\overline{h} : \mathcal{X}^{(R)} \to I$ and defined by

$$\forall \bar{\mathbf{x}} \in \mathcal{X}^{(R)}, \quad \overline{h}(\bar{\mathbf{x}}) \triangleq \psi(h(x[1]), \ldots, h(x[r])).$$

The hypothesis class for bags given $\mathcal{H}$ and $\psi$ is denoted $\overline{\mathcal{H}} \triangleq \{\overline{h} \mid h \in \mathcal{H}\}$. Importantly, the identity of $\psi$ is known to the learner a-priori, thus each $\psi$ defines a different generalized MIL problem. For instance, in classical MIL, $I = \{-1, +1\}$ and $\psi$ is the Boolean OR.

We assume the labeled bags are drawn from a fixed distribution $D$ over $\mathcal{X}^{(R)} \times \{-1, +1\}$, where each pair drawn from $D$ constitutes a bag and its binary label. Given a range $I \subseteq \mathbb{R}$ of possible label predictions, we define a loss function $\ell : \{-1, +1\} \times I \to \mathbb{R}$, where $\ell(y, \hat{y})$ is the loss incurred if the true label is $y$ and the predicted label is $\hat{y}$. The true loss of a bag-classifier $h : \mathcal{X}^{(R)} \to I$ is denoted by $\ell(h, D) \triangleq \mathbb{E}_{(\bar{\mathbf{X}}, Y) \sim D}[\ell(Y, h(\bar{\mathbf{X}}))]$. We say that a sample or a distribution are *realizable* by $\overline{\mathcal{H}}$ if there is a hypothesis $h \in \overline{\mathcal{H}}$ that classifies them with zero loss.

The MIL learner receives a labeled sample of bags $\{(\bar{\mathbf{x}}_1, y_1), \ldots, (\bar{\mathbf{x}}_m, y_m)\} \subseteq \mathcal{X}^{(R)} \times \{-1, +1\}$ drawn from $D^m$, and returns a classifier $\hat{h} : \mathcal{X}^{(R)} \to I$. The goal of the learner is to return $\hat{h}$ that has a low loss $\ell(\hat{h}, D)$ compared to the minimal loss that can be achieved with the bag hypothesis class, denoted by $\ell^*(\overline{\mathcal{H}}, D) \triangleq \inf_{h\in\overline{\mathcal{H}}} \ell(h, D)$. The empirical loss of a classifier for bags on a labeled sample $S$ is $\ell(h, S) \triangleq \mathbb{E}_{(X,Y)\sim S}[\ell(Y, h(\bar{\mathbf{X}}))]$. For an unlabeled set of bags $S = \{\bar{\mathbf{x}}_i\}_{i\in[m]}$, we denote the multi-set of instances in the bags of $S$ by $S^\cup \triangleq \{x_i[j] \mid i \in [m], j \in [|\bar{\mathbf{x}}_i|]\}$. Since this is a multi-set, any instance which repeats in several bags in $S$ is represented the same amount of time in $S^\cup$.

**Classes of Real-Valued bag-functions**

In classical MIL the bag function is the Boolean OR over binary labels, that is $I = \{-1, +1\}$ and $\psi = \mathrm{OR} : \{-1, +1\}^{(R)} \to \{-1, +1\}$. A natural extension of the Boolean OR to a function over reals is the $\max$ function. We further consider two classes of bag functions over reals, each representing a different generalization of the $\max$ function, which conserves a different subset of its properties.

The first class we consider is the class of bag-functions that extend monotone Boolean functions. Monotone Boolean functions map Boolean vectors to $\{-1, +1\}$, such that the map is monotone-increasing in each of the inputs. The set of monotone Boolean functions is exactly the set of functions that can be represented by some composition of AND and OR functions, thus it includes the Boolean OR. The natural extension of monotone Boolean functions to real functions over real vectors is achieved by replacing OR with $\max$ and AND with $\min$. Formally, we define extensions of monotone Boolean functions as follows.

**Definition 2** *A function from $\mathbb{R}^n$ into $\mathbb{R}$ is* an extension of an $n$-ary monotone Boolean function *if it belongs to the set $\mathcal{M}_n$ defined inductively as follows, where the input to a function is $\mathbf{z} \in \mathbb{R}^n$:*

$$\begin{array}{ll} (1)\ \forall j \in [n], & \mathbf{z} \mapsto z[j] \in \mathcal{M}_n; \\ (2)\ \forall k \in \mathbb{N}^+, & f_1, \ldots, f_k \in \mathcal{M}_n \Longrightarrow \mathbf{z} \mapsto \max_{j\in[k]}\{f_j(\mathbf{z})\} \in \mathcal{M}_n; \\ (3)\ \forall k \in \mathbb{N}^+, & f_1, \ldots, f_k \in \mathcal{M}_n \Longrightarrow \mathbf{z} \mapsto \min_{j\in[k]}\{f_j(\mathbf{z})\} \in \mathcal{M}_n. \end{array}$$

*We say that a bag-function $\psi : \mathbb{R}^{(R)} \to \mathbb{R}$ extends monotone Boolean functions if for all $n \in R$, $\psi_{|\mathbb{R}^n} \in \mathcal{M}_n$.*

The class of extensions to Boolean functions thus generalizes the $\max$ function in a natural way.

The second class of bag functions we consider generalizes the $\max$ function by noting that for bounded inputs, the $\max$ function can be seen as a variant of the infinity-norm $\|\mathbf{z}\|_\infty = \max |z[i]|$. Another natural bag-function over reals is the average function, defined as $\psi(\mathbf{z}) = \frac{1}{n}\sum_{i\in[n]} z_i$, which can be seen as a variant of the 1-norm $\|\mathbf{z}\|_1 = \sum_{i\in[n]} |z[i]|$. More generally, we treat the case where the hypotheses map into $I = [-1, 1]$, and consider the class of bag functions inspired by a $p$-norm, defined as follows.

**Definition 3** *For $p \in [1, \infty)$, the* $p$-norm bag function $\psi_p : [-1, +1]^{(R)} \to [-1, +1]$ *is defined by:*

$$\forall \mathbf{z} \in \mathbb{R}^n, \quad \psi_p(\mathbf{z}) \triangleq \left(\frac{1}{n}\sum_{i=1}^{n}(z[i] + 1)^p\right)^{1/p} - 1.$$

*For $p = \infty$, Define $\psi_\infty \equiv \lim_{p\to\infty} \psi_p$.*

Since the inputs of $\psi_p$ are in $[-1,+1]$, we have $\psi_p(\mathbf{z}) \equiv n^{-1/p} \cdot \|\mathbf{z}+\mathbf{1}\|_p - 1$ where $n$ is the length of $\mathbf{z}$. Note that the average function is simply $\psi_1$, and $\psi_\infty \equiv \|\mathbf{z}+\mathbf{1}\|_\infty - 1 \equiv \max$. Other values of $p$ fall between these two extremes: Due to the $p$-norm inequality, which states that for all $p \in [1,\infty)$ and $\mathbf{x} \in \mathbb{R}^n$, $\frac{1}{n}\|\mathbf{x}\|_1 \leq n^{-1/p}\|\mathbf{x}\|_p \leq \|\mathbf{x}\|_\infty$, we have that for all $\mathbf{z} \in [-1,+1]^n$

$$\text{average} \equiv \psi_1(\mathbf{z}) \leq \psi_p(\mathbf{z}) \leq \psi_\infty(\mathbf{z}) \equiv \max. \tag{1}$$

Many of our results hold when the scale of the output of the bag-function is related to the scale of its inputs. Formally, we consider cases where the output of the bag-function does not change by much unless its inputs change by much. This is formalized in the following definition of a Lipschitz bag function.

**Definition 4** *A bag function $\psi : \mathbb{R}^{(R)} \to \mathbb{R}$ is $c$-*Lipschitz with respect to the infinity norm *for $c > 0$ if*

$$\forall n \in R, \forall \mathbf{a}, \mathbf{b} \in \mathbb{R}^n, \quad |\psi(\mathbf{a}) - \psi(\mathbf{b})| \leq c\|\mathbf{a}-\mathbf{b}\|_\infty.$$

The average bag-function and the $\max$ bag functions are 1-Lipschitz. Moreover, all extensions of monotone Boolean functions are 1-Lipschitz with respect to the infinity norm—this is easy to verify by induction on Def. 2. All $p$-norm bag functions are also 1-Lipschitz, as the following derivation shows:

$$|\psi_p(\mathbf{a}) - \psi_p(\mathbf{b})| = n^{-1/p} \cdot |\, \|\mathbf{a}+1\|_p - \|\mathbf{b}+1\|_p| \leq n^{-1/p} \cdot \|\mathbf{a}-\mathbf{b}\|_p \leq \|\mathbf{a}-\mathbf{b}\|_\infty.$$

Thus, our results for Lipschitz bag-functions hold in particular for the two bag-function classes we have defined here, and in specifically for the $\max$ function.

## 3. Binary MIL

In this section we consider binary MIL. In binary MIL we let $I = \{-1,+1\}$, thus we have a binary instance hypothesis class $\mathcal{H} \subseteq \{-1,+1\}^{\mathcal{X}}$. We further let our loss be the zero-one loss, defined by $\ell_{0/1}(y,\hat{y}) = \mathbb{1}[y \neq \hat{y}]$. The distribution-free sample complexity of learning relative to a binary hypothesis class with the zero-one loss is governed by the VC-dimension of the hypothesis class (Vapnik and Chervonenkis, 1971). Thus we bound the VC-dimension of $\overline{\mathcal{H}}$ as a function of the maximal possible bag size $r = \max R$, and of the VC-dimension of $\mathcal{H}$. We show that the VC-dimension of $\overline{\mathcal{H}}$ is at most logarithmic in $r$, and at most linear in the VC-dimension of $\mathcal{H}$, for any bag-labeling function $\psi : \{-1,+1\}^{(R)} \to \{-1,+1\}$. It follows that the sample complexity of MIL grows only logarithmically with the size of the bag. Thus MIL is feasible even for quite large bags. In fact, based on the results we show henceforth, Sabato et al. (2010) have shown that MIL can sometimes be used to accelerate even single-instance learning. We further provide lower bounds that show that the dependence of the upper bound on $r$ and on the VC-dimension of $\mathcal{H}$ is imperative, for a large class of Boolean bag-labeling functions. We also show a matching lower bound for the VC-dimension of classical MIL with separating hyperplanes.

### 3.1 VC-Dimension Upper Bound

Our first theorem establishes a VC-Dimension upper bound for generalized MIL. To prove the theorem we require the following useful lemma.

**Lemma 5** *For any $R \subseteq \mathbb{N}$ and any bag function $\psi : \{-1,+1\}^{(R)} \to \{-1,+1\}$, and for any hypothesis class $\mathcal{H} \subseteq \{-1,+1\}^{\mathcal{X}}$ and a finite set of bags $S \subseteq \mathcal{X}^{(R)}$,*

$$\left|\overline{\mathcal{H}}_{|S}\right| \leq \left|\mathcal{H}_{|S^{\cup}}\right|.$$

**Proof** Let $h_1, h_2 \in \overline{\mathcal{H}}$ be bag hypotheses. There exist instance hypotheses $g_1, g_2 \in \mathcal{H}$ such that $\overline{g}_i = h_i$ for $i = 1, 2$. Assume that $h_{1|S} \neq h_{2|S}$. We show that $g_{1|S^{\cup}} \neq g_{2|S^{\cup}}$, thus proving the lemma.

From the assumption it follows that $\overline{g}_{1|S} \neq \overline{g}_{2|S}$. Thus there exists at least one bag $\mathbf{x} \in S$ such that $\overline{g}_2(\mathbf{x}) \neq \overline{g}_2(\mathbf{x})$. Denote its size by $n$. We have $\psi(g_1(x[1]), \ldots, g_1(x[n])) \neq \psi(g_2(x[1]), \ldots, g_2(x[n]))$. Hence there exists a $j \in [n]$ such that $g_1(x[j]) \neq g_2(x[j])$. By the definition of $S^{\cup}$, $x[j] \in S^{\cup}$. Therefore $g_{1|S^{\cup}} \neq g_{2|S^{\cup}}$. ■

**Theorem 6** *Assume that $\mathcal{H}$ is a hypothesis class with a finite VC-dimension $d$. Let $r \in \mathbb{N}$ and assume that $R \subseteq [r]$. Let the bag-labeling function $\psi : \{-1,+1\}^{(R)} \to \{-1,+1\}$ be some Boolean function. Denote the VC-dimension of $\overline{\mathcal{H}}$ by $d_r$. We have*

$$d_r \leq \max\{16, 2d\log(2er)\}.$$

**Proof** For a set of hypotheses $\mathcal{J}$, denote by $\mathcal{J}_{|A}$ the restriction of each of its members to $A$, so that $\mathcal{J}_A \triangleq \{h_{|A} \mid h \in \mathcal{J}\}$. Since $d_r$ is the VC-dimension of $\overline{\mathcal{H}}$, there exists a set of bags $S \subseteq \mathcal{X}^{(R)}$ of size $d_r$ that is shattered by $\overline{\mathcal{H}}$, so that $|\overline{\mathcal{H}}_{|S}| = 2^{d_r}$. By Lemma 5 $|\overline{\mathcal{H}}_{|S}| \leq |\mathcal{H}_{|S^{\cup}}|$, therefore $2^{d_r} \leq |\mathcal{H}_{|S^{\cup}}|$. In addition, $R \subseteq [r]$ implies $|S^{\cup}| \leq rd_r$. By applying Sauer's lemma (Sauer, 1972; Vapnik and Chervonenkis, 1971) to $\mathcal{H}$ we get

$$2^{d_r} \leq |\mathcal{H}_{|S^{\cup}}| \leq \left(\frac{e|S^{\cup}|}{d}\right)^d \leq \left(\frac{erd_r}{d}\right)^d,$$

Where $e$ is the base of the natural logarithm. It follows that $d_r \leq d(\log(er) - \log d) + d\log d_r$. To provide an explicit bound for $d_r$, we bound $d\log d_r$ by dividing to cases:

1. Either $d\log d_r \leq \frac{1}{2}d_r$, thus $d_r \leq 2d(\log(er) - \log d) \leq 2d\log(er)$,
2. or $\frac{1}{2}d_r < d\log d_r$. In this case,
   (a) either $d_r \leq 16$,
   (b) or $d_r > 16$. In this case $\sqrt{d_r} < d_r/\log d_r < 2d$, thus $d\log d_r = 2d\log\sqrt{d_r} \leq 2d\log 2d$. Substituting in the implicit bound we get $d_r \leq d(\log(er) - \log d) + 2d\log 2d \leq 2d\log(2er)$.

Combining the cases we have $d_r \leq \max\{16, 2d\log(2er)\}$. ■

### 3.2 VC-Dimension Lower Bounds

In this section we show lower bounds for the VC-dimension of binary MIL, indicating that the dependence on $d$ and $r$ in Theorem 6 is tight in two important settings.

We say that a bag-function $\psi : \{-1,+1\}^{(R)} \rightarrow \{-1,+1\}$ is $r$-*sensitive* if there exists a number $n \in R$ and a vector $\mathbf{c} \in \{-1,+1\}^n$ such that for at least $r$ different numbers $j_1, \ldots, j_r \in [n]$, $\psi(c[1], \ldots, c[j_i], \ldots, c[n]) \neq \psi(c[1], \ldots, -c[j_i], \ldots, c[n])$. Many commonly used Boolean functions, such as OR, AND, Parity, and all their variants that stem from negating some of the inputs, are $r$-sensitive for every $r \in R$. Our first lower bound shows if $\psi$ is $r$-sensitive, the bound in Theorem 6 cannot be improved without restricting the set of considered instance hypothesis classes.

**Theorem 7** *Assume that the bag function $\psi : \{-1,+1\}^{(R)} \rightarrow \{-1,+1\}$ is $r$-sensitive for some $r \in \mathbb{N}$. For any natural $d$ and any instance domain $\mathcal{X}$ with $|\mathcal{X}| \geq rd\lfloor\log(r)\rfloor$, there exists a hypothesis class $\mathcal{H}$ with a VC-dimension at most $d$, such that the VC dimension of $\overline{\mathcal{H}}$ is at least $d\lfloor\log(r)\rfloor$.*

**Proof** Since $\psi$ is $r$-sensitive, there are a vector $\mathbf{c} \in \{-1,+1\}^n$ and a set $J \subseteq n$ such that $|J| = r$ and $\forall j \in J, \psi(c[1], \ldots, c[n]) \neq \psi(c[1], \ldots, -c[j], \ldots, c[n])$. Since $\psi$ maps all inputs to $\{-1,+1\}$, it follows that $\forall j \in J, \psi(c[1], \ldots, -c[j], \ldots, c[n]) = -\psi(c[1], \ldots, c[n])$. Denote $a = \psi(c[1], \ldots, c[n])$. Then we have

$$\forall j \in J, y \in \{-1,+1\}, \quad \psi(c[1], \ldots, c[j] \cdot y, \ldots, c[n]) = a \cdot y. \tag{2}$$

For simplicity of notation, we henceforth assume w.l.o.g. that $n = r$ and $J = [r]$.

Let $S \subseteq \mathcal{X}^r$ be a set of $d\lfloor\log(r)\rfloor$ bags of size $r$, such that all the instances in all the bags are distinct elements of $\mathcal{X}$. Divide $S$ into $d$ mutually exclusive subsets, each with $\lfloor\log(r)\rfloor$ bags. Denote bag $p$ in subset $t$ by $\bar{\mathbf{x}}_{(p,t)}$. We define the hypothesis class

$$\mathcal{H} \triangleq \{h[k_1, \ldots, k_d] \mid \forall i \in [d], k_i \in [2^{\lfloor\log(r)\rfloor}]\},$$

where $h[k_1, \ldots, k_d]$ is defined as follows (see illustration in Table 1): For $x \in \mathcal{X}$ which is not an instance of any bag in $S$, $h[k_1, \ldots, k_d] = -1$. For $x = x_{(p,t)}[j]$, let $b_{(p,n)}$ be bit $p$ in the binary representation of the number $n$, and define

$$h[k_1, \ldots, k_d](x_{(p,t)}[j]) = \begin{cases} c[j] \cdot a(2b_{(p,j-1)} - 1) & j = k_t, \\ c[j] & j \neq k_t. \end{cases}$$

We now show that $S$ is shattered by $\overline{\mathcal{H}}$, indicating that the VC-dimension of $\overline{\mathcal{H}}$ is at least $|S| = d\lfloor\log(r)\rfloor$. To complete the proof, we further show that the VC-dimension of $\mathcal{H}$ is no more than $d$.

**$S$ is shattered by $\overline{\mathcal{H}}$:** Let $\{y_{(p,t)}\}_{p \in \lfloor\log(r)\rfloor, t \in [d]}$ be some labeling over $\{-1,+1\}$ for the bags in $S$. For each $t \in [d]$ let

$$k_t \triangleq 1 + \sum_{p=1}^{\lfloor\log(r)\rfloor} \frac{y_{(p,t)} + 1}{2} \cdot 2^{p-1}.$$

Then by Eq. (2), for all $p \in [\lfloor\log(r)\rfloor]$ and $t \in [d]$,

$$\begin{aligned} \overline{h}[k_1, \ldots, k_d](\bar{\mathbf{x}}_{(p,t)}) &= \psi(c[1], \ldots, c[k_t] \cdot a(2b_{(p,k_t-1)} - 1), \ldots, c[r]) \\ &= a^2(2b_{(p,k_t-1)} - 1) = 2b_{(p,k_t-1)} - 1 = y_{(p,t)}. \end{aligned}$$

Thus $h[k_1, \ldots, k_d]$ labels $S$ according to $\{y_{(p,t)}\}$.

| $t$ | $p$ | Instance label $h(x_{(p,t)}[r])$ | | | | | | | | Bag label $\overline{h}(\bar{\mathbf{x}}_i)$ |
|---|---|---|---|---|---|---|---|---|---|---|
| 1 | 1 | $-$ | $-$ | $-$ | $+$ | $-$ | $-$ | $-$ | $-$ | $+$ |
| | 2 | $-$ | $-$ | $-$ | $+$ | $-$ | $-$ | $-$ | $-$ | $+$ |
| | 3 | $-$ | $-$ | $-$ | $-$ | $-$ | $-$ | $-$ | $-$ | $-$ |
| 2 | 1 | $-$ | $-$ | $-$ | $-$ | $-$ | $-$ | $-$ | $+$ | $+$ |
| | 2 | $-$ | $-$ | $-$ | $-$ | $-$ | $-$ | $-$ | $+$ | $+$ |
| | 3 | $-$ | $-$ | $-$ | $-$ | $-$ | $-$ | $-$ | $+$ | $+$ |
| 3 | 1 | $-$ | $-$ | $-$ | $-$ | $-$ | $-$ | $-$ | $-$ | $-$ |
| | 2 | $-$ | $+$ | $-$ | $-$ | $-$ | $-$ | $-$ | $-$ | $+$ |
| | 3 | $-$ | $-$ | $-$ | $-$ | $-$ | $-$ | $-$ | $-$ | $-$ |

Table 1: An example of the hypotheses $h = h[4, 8, 3]$, with $\psi = \text{OR}$ (so that $\mathbf{c}$ is the all $-1$ vector), $r = 8$, and $d = 3$. Each line represents a bag in $S$, each column represents an instance in the bag.

**The VC-dimension of $\mathcal{H}$ is no more than $d$:** Let $A \subseteq \mathcal{X}$ of size $d + 1$. If there is an element in $A$ which is not an instance in $S$ then this element is labeled $-1$ by all $h \in \mathcal{H}$, therefore $A$ is not shattered. Otherwise, all elements in $A$ are instances in bags in $S$. Since there are $d$ subsets of $S$, there exist two elements in $A$ which are instances of bags in the same subset $t$. Denote these instances by $x(p_1, t)[j_1]$ and $x(p_2, t)[j_2]$. Consider all the possible labelings of the two elements by hypotheses in $\mathcal{H}$. If $A$ is shattered, there must be four possible labelings for these elements. However, by the definition of $h[k_1, \ldots, k_d]$ it is easy to see that if $j_1 = j_2 = j$ then there are at most two possible labelings by hypotheses in $\mathcal{H}$, and if $j_1 \neq j_2$ then there are at most three possible labelings. Thus $A$ is not shattered by $\mathcal{H}$, hence the VC-dimension of $\mathcal{H}$ is no more than $d$. ■

Theorem 10 below provides a lower bound for the VC-dimension of MIL for the important case where the bag-function is the Boolean OR and the hypothesis class is the class of separating hyperplanes in $\mathbb{R}^n$. For $\mathbf{w} \in \mathbb{R}^n$, the function $h_{\mathbf{w}} : \mathbb{R}^n \to \{-1, +1\}$ is defined by $h_{\mathbf{w}}(\mathbf{x}) = \operatorname{sign}(\langle \mathbf{w}, \mathbf{x} \rangle)$. The hypothesis class of linear classifiers is $\mathcal{W}_n \triangleq \{h_{\mathbf{w}} \mid \mathbf{w} \in \mathbb{R}^n\}$. Let $r \in \mathbb{N}$. We denote the VC-dimension of $\overline{\mathcal{W}}_n$ for $R = \{r\}$ and $\psi = \text{OR}$ by $d_{r,n}$. We prove a lower bound for $d_{r,n}$ using two lemmas: Lemma 8 provides a lower bound for $d_{r,3}$, and Lemma 9 links $d_{r,n}$ for small $n$ with $d_{r,n}$ for large $n$. The resulting general lower bound, which holds for $r = \max R$, is then stated in Theorem 10.

**Lemma 8** *Let $d_{r,n}$ be the VC-dimension of $\overline{\mathcal{W}}_n$ as defined above. Then $d_{r,3} \geq \lfloor \log(2r) \rfloor$.*

**Proof** Denote $L \triangleq \lfloor \log(2r) \rfloor$. We will construct a set $S$ of $L$ bags of size $r$ that is shattered by $\mathcal{W}_3$. The construction is illustrated in Figure 1.

Let $\mathbf{n} = (n_1, \ldots, n_K)$ be a sequence of indices from $[L]$, created by concatenating all the subsets of $[L]$ in some arbitrary order, so that $K = L2^{L-1}$, and every index appears $2^{L-1} \leq r$ times in $\mathbf{n}$. Define a set $A = \{\mathbf{a}_k \mid k \in [K]\} \subseteq \mathbb{R}^3$ where $\mathbf{a}_k \triangleq (\cos(2\pi k/K), \sin(2\pi k/K), 1) \in \mathbb{R}^3$, so that $\mathbf{a}_1, \ldots, \mathbf{a}_K$ are equidistant on a unit circle on a plane embedded in $\mathbb{R}^3$. Define the set of bags $S = \{\bar{\mathbf{x}}_1, \ldots, \bar{\mathbf{x}}_L\}$ such that $\bar{\mathbf{x}}_i = (x_i[1], \ldots, x_i[r])$ where $\{x_i[j] \mid j \in [r]\} = \{a_k \mid n_k = i\}$.

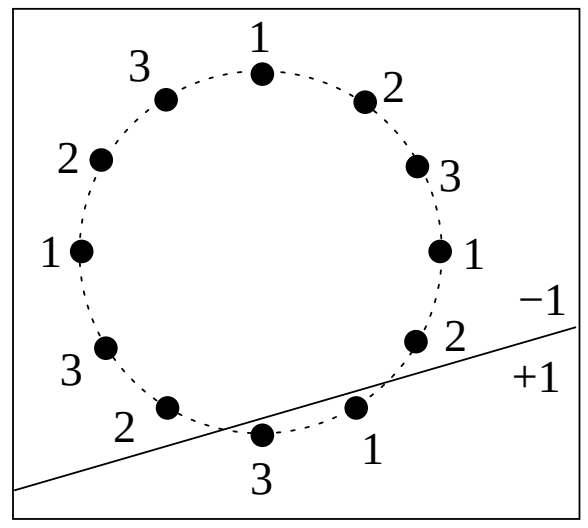

Figure 1: An illustration of the constructed shattered set, with $r = 4$ and $L = \log 4 + 1 = 3$. Each dot corresponds to an instance. The numbers next to the instances denote the bag to which an instance belongs, and match the sequence $N$ defined in the proof. In this illustration bags 1 and 3 are labeled as positive by the bag-hypothesis represented by the solid line.

We now show that $S$ is shattered by $\mathcal{W}_3$: Let $(y_1, \ldots, y_L)$ be some binary labeling of $L$ bags, and let $Y = \{i \mid y_i = +1\}$. By the definition of $\mathbf{n}$, there exist $j_1, j_2$ such that $Y = \{n_k \mid j_1 \leq k \leq j_2\}$. Clearly, there exists a hyperplane $\mathbf{w} \in \mathbb{R}^3$ that separates the vectors $\{\mathbf{a}_k \mid j_1 \leq k \leq j_2\}$ from the rest of the vectors in $A$. Thus $\text{sign}(\langle \mathbf{w}, \mathbf{a}_k \rangle) = +1$ if and only if $j_1 \leq k \leq j_2$. It follows that $\overline{h}_{\mathbf{w}}(\bar{\mathbf{x}}_i) = +1$ if and only if there is a $k \in \{j_1, \ldots, j_2\}$ such that $\mathbf{a}_k$ is an instance in $\bar{\mathbf{x}}_i$, that is such that $n_k = i$. This condition holds if and only if $i \in Y$, hence $\overline{h}_{\mathbf{w}}$ classifies $S$ according to the given labeling. It follows that $S$ is shattered by $\mathcal{W}_3$, therefore $d_{r,3} \geq |S| = \lfloor \log(2r) \rfloor$. ■

**Lemma 9** *Let $k, n, r$ be natural number such that $k \leq n$. Then $d_{r,n} \geq \lfloor n/k \rfloor d_{r,k}$.*

**Proof** For a vector $\mathbf{x} \in \mathbb{R}^k$ and a number $t \in \{0, \ldots, \lfloor n/k \rfloor\}$ define the vector $s(\mathbf{x}, t) \triangleq (0, \ldots, 0, x[1], \ldots, x[k], 0, \ldots, 0) \in \mathbb{R}^n$, where $x[1]$ is at coordinate $kt + 1$. Similarly, for a bag $\bar{\mathbf{x}}_i = (\mathbf{x}_i[1], \ldots, \mathbf{x}_i[r]) \in (\mathbb{R}^k)^r$, define the bag $s(\bar{\mathbf{x}}_i, t) \triangleq (s(\mathbf{x}_i[1], t), \ldots, s(\mathbf{x}_i[r], t)) \in (\mathbb{R}^n)^r$.

Let $S_k = \{\bar{\mathbf{x}}_i\}_{i \in [d_{r,k}]} \subseteq (\mathbb{R}^k)^r$ be a set of bags with instances in $\mathbb{R}^k$ that is shattered by $\overline{\mathcal{W}}_k$. Define $S_n$, a set of bags with instances in $\mathbb{R}^n$: $S_n \triangleq \{s(\bar{\mathbf{x}}_i, t)]\}_{i \in [d_{r,k}], t \in [\lfloor n/k \rfloor]} \subseteq (\mathbb{R}^n)^r$. Then $S_n$ is shattered by $\mathcal{W}_n$: Let $\{y_{(i,t)}\}_{i \in [d_{r,k}], t \in [\lfloor n/k \rfloor]}$ be some labeling for $S_n$. $S_k$ is shattered by $\mathcal{W}_k$, hence there are separators $\mathbf{w}_1, \ldots, \mathbf{w}_{\lfloor n/k \rfloor} \in \mathbb{R}^k$ such that $\forall i \in [d_{r,k}], t \in \lfloor n/k \rfloor, \quad \overline{h}_{\mathbf{w}_t}(\bar{\mathbf{x}}_i) = y_{(i,t)}$.

Set $\mathbf{w} \triangleq \sum_{t=0}^{\lfloor n/k \rfloor} s(\mathbf{w}_t, t)$. Then $\langle \mathbf{w}, s(\mathbf{x}, t) \rangle = \langle \mathbf{w}_t, \mathbf{x} \rangle$. Therefore

$$\begin{aligned}\overline{h}_{\mathbf{w}}(s(\bar{\mathbf{x}}_i, t)) &= \text{OR}(\text{sign}(\langle \mathbf{w}, s(\mathbf{x}_i[1], t) \rangle), \ldots, \text{sign}(\langle \mathbf{w}, s(\mathbf{x}_i[r], t) \rangle)) \\ &= \text{OR}(\text{sign}(\langle \mathbf{w}_t, \mathbf{x}_i[1] \rangle), \ldots, \text{sign}(\langle \mathbf{w}_t, \mathbf{x}_i[r] \rangle)) = \overline{h}_{\mathbf{w}_t}(\bar{\mathbf{x}}_i) = y_{(i,t)}.\end{aligned}$$

$S_n$ is thus shattered, hence $d_{r,n} \geq |S_n| = \lfloor n/k \rfloor d_{r,k}$. ■

The desired theorem is an immediate consequence of the two lemmas above, by noting that whenever $r \in R$, the VC-dimension of $\overline{\mathcal{W}}_n$ is at least $d_{r,n}$.

**Theorem 10** *Let $\mathcal{W}_n$ be the class of separating hyperplanes in $\mathbb{R}^n$ as defined above. Assume that the bag function is $\psi = OR$ and the set of allowed bag sizes is $R$. Let $r = \max R$. Then the VC-dimension of $\overline{\mathcal{W}}_n$ is at least $\lfloor n/3 \rfloor \lfloor \log 2r \rfloor$.*

### 3.3 Pseudo-dimension for thresholded functions

In this section we consider binary hypothesis classes that are generated from real-valued functions using thresholds. Let $\mathcal{F} \subseteq \mathbb{R}^{\mathcal{X}}$ be a set of real valued functions. The binary hypothesis class of thresholded functions generated by $\mathcal{F}$ is $T_{\mathcal{F}} = \{(x, z) \mapsto \mathrm{sign}(f(x) - z) \mid f \in \mathcal{F}, z \in \mathbb{R}\}$, where $x \in \mathcal{X}$ and $z \in \mathbb{R}$. The sample complexity of learning with $T_{\mathcal{F}}$ and the zero-one loss is governed by the pseudo-dimension of $\mathcal{F}$, which is equal to the VC-dimension of $T_{\mathcal{F}}$ (Pollard, 1984). In this section we consider a bag-labeling function $\psi : \mathbb{R}^{(R)} \to \mathbb{R}$, and bound the pseudo-dimension of $\overline{\mathcal{F}}$, thus providing an upper bound on the sample complexity of binary MIL with $T_{\overline{\mathcal{F}}}$. The following bound holds for bag-labeling functions that extend monotone Boolean functions, defined in Def. 2.

**Theorem 11** *Let $\mathcal{F} \subseteq \mathbb{R}^{\mathcal{X}}$ be a function class with pseudo-dimension $d$ . Let $R \subseteq [r]$, and assume that $\psi : \mathbb{R}^{(R)} \to \mathbb{R}$ extends monotone Boolean functions. Let $d_r$ be the pseudo-dimension of $\overline{\mathcal{F}}$. Then*

$$d_r \leq \max\{16, 2d \log(2er)\}.$$

**Proof** First, by Def. 2, we have that for any $\psi$ which extends monotone Boolean functions, any $n \in R$ and any $\mathbf{y} \in \mathbb{R}^n$,

$$\begin{aligned}\mathrm{sign}(\psi(y[1], \dots, y[n]) - z) &= \mathrm{sign}(\psi(y[1] - z, \dots, y[n] - z)) \\ &= \psi(\mathrm{sign}(y[1] - z, \dots, y[n] - z)).\end{aligned} \tag{3}$$

This can be seen by noting that each of the equalities holds for each of the operations allowed by $\mathcal{M}_n$ for each $n$, thus by induction they hold for all functions in $\mathcal{M}_n$ and all combinations of them.

For a real-valued function $f$ let $t_f : \mathcal{X} \times \mathbb{R} \to \{-1, +1\}$ be defined by $t_f(y, z) = \mathrm{sign}(f(y) - z)$. We have $T_{\mathcal{F}} = \{t_f \mid f \in \mathcal{F}\}$, and $T_{\overline{\mathcal{F}}} = \{t_{\overline{f}} \mid f \in \mathcal{F}\}$. In addition, for all $f \in \mathcal{F}$, $z \in \mathbb{R}$, $n \in R$ and $\bar{\mathbf{x}} \in \mathcal{X}^n$, we have

$$\begin{aligned}t_{\overline{f}}(\bar{\mathbf{x}}, z) &= \mathrm{sign}(\overline{f}(\bar{\mathbf{x}}) - z) = \mathrm{sign}(\psi(f(x[1]), \dots, f(x[n])) - z) \\ &= \psi(\mathrm{sign}(f(x[1]) - z, \dots, f(x[n]) - z)) \\ &= \psi(t_f(x[1], z), \dots, t_f(x[n], z)) = \overline{t_f}(\bar{\mathbf{x}}, z),\end{aligned} \tag{4}$$

where the equality on line (4) follows from Eq. (3). Therefore

$$T_{\overline{\mathcal{F}}} = \{t_{\overline{f}} \mid f \in \mathcal{F}\} = \{\overline{t_f} \mid f \in \mathcal{F}\} = \{\overline{h} \mid h \in T_{\mathcal{F}}\} = \overline{T_{\mathcal{F}}}.$$

The VC-dimension of $T_{\mathcal{F}}$ is equal to the pseudo-dimension of $\mathcal{F}$, which is $d$. Thus, by Theorem 6 and the equality above, the VC-dimension of $T_{\overline{\mathcal{F}}}$ is bounded by $\max\{16, 2d \log(2er)\}$. The proof is completed by noting that $d_r$, the pseudo-dimension of $\overline{\mathcal{F}}$, is exactly the VC-dimension of $T_{\overline{\mathcal{F}}}$. ■

This concludes our results for distribution-free sample complexity of Binary MIL. In Section 6 we provide sample complexity analysis for distribution-dependent binary MIL, as a function of the average bag size.

## 4. Covering Numbers bounds for MIL

Covering numbers are a useful measure of the complexity of a function class, since they allow bounding the sample complexity of a class in various settings, based on uniform convergence guarantees (see e.g. Anthony and Bartlett, 1999). In this section we provide a lemma that relates the covering numbers of bag hypothesis classes with those of the underlying instance hypothesis class. We will use this lemma in subsequent sections to derive sample complexity upper bounds for additional settings of MIL. Let $\mathcal{F} \subseteq \mathbb{R}^A$ be a set of real-valued functions over some domain $A$. A $\gamma$-cover of $\mathcal{F}$ with respect to a norm $\|\cdot\|_\circ$ defined on functions is a set of functions $\mathcal{C} \subseteq \mathbb{R}^A$ such that for any $f \in \mathcal{F}$ there exists a $g \in \mathcal{C}$ such that $\|f - g\|_\circ \leq \gamma$. The *covering number* for given $\gamma > 0$, $\mathcal{F}$ and $\circ$, denoted by $\mathcal{N}(\gamma, \mathcal{F}, \circ)$, is the size of the smallest such $\gamma$-covering for $\mathcal{F}$.

Let $S \subseteq A$ be a finite set. We consider coverings with respect to the $L_p(S)$ norm for $p \geq 1$, defined by

$$\|f\|_{L_p(S)} \triangleq \left(\frac{1}{|S|}\sum_{s\in S}|f(s)|^p\right)^{1/p}.$$

For $p = \infty$, $L_\infty(S)$ is defined by $\|f\|_{L_\infty(S)} \triangleq \max_{s\in S}|f(S)|$. The covering number of $\mathcal{F}$ for a sample size $m$ with respect to the $L_p$ norm is

$$\mathcal{N}_m(\gamma, \mathcal{F}, p) \triangleq \sup_{S\subseteq A:|S|=m} \mathcal{N}(\gamma, \mathcal{F}, L_p(S)).$$

A small covering number for a function class implies faster uniform convergence rates, hence smaller sample complexity for learning. The following lemma bounds the covering number of bag hypothesis-classes whenever the bag function is Lipschitz with respect to the infinity norm (see Def. 4). Recall that all extensions of monotone Boolean functions (Def. 2) and all $p$-norm bag-functions (Def. 3) are 1-Lipschitz, thus the following lemma holds for them with $a = 1$.

**Lemma 12** *Let $R \subseteq \mathbb{N}$ and suppose the bag function $\psi : \mathbb{R}^{(R)} \to \mathbb{R}$ is $a$-Lipschitz with respect to the infinity norm, for some $a > 0$. Let $S \subseteq \mathcal{X}^{(R)}$ be a finite set of bags, and let $r$ be the average size of a bag in $S$. For any $\gamma > 0$, $p \in [1, \infty]$, and hypothesis class $\mathcal{H} \subseteq \mathbb{R}^{\mathcal{X}}$,*

$$\mathcal{N}(\gamma, \overline{\mathcal{H}}, L_p(S)) \leq \mathcal{N}(\frac{\gamma}{ar^{1/p}}, \mathcal{H}, L_p(S^\cup)).$$

**Proof** First, note that by the Lipschitz condition on $\psi$, for any bag $\bar{\mathbf{x}}$ of size $n$ and hypotheses $h, g \in \mathcal{H}$,

$$|\overline{h}(\bar{\mathbf{x}}) - \overline{g}(\bar{\mathbf{x}})| = |\psi(h(x[1]), \ldots, h(x[n])) - \psi(g(x[1]), \ldots, g(x[n]))| \leq a \max_{x\in\bar{\mathbf{x}}} |h(x) - g(x)|. \quad (5)$$

Let $\mathcal{C}$ be a minimal $\gamma$-cover of $\mathcal{H}$ with respect to the norm defined by $L_p(S^\cup)$, so that $|\mathcal{C}| = \mathcal{N}(\gamma, \mathcal{H}, L_p(S^\cup))$. For every $h \in \mathcal{H}$ there exists a $g \in \mathcal{C}$ such that $\|h - g\|_{L_p(S^\cup)} \leq \gamma$. Assume

$p < \infty$. Then by Eq. (5)

$$\begin{aligned}
\|\overline{h} - \overline{g}\|_{L_p(S)} &= \left(\frac{1}{|S|}\sum_{\bar{\mathbf{x}} \in S} |\overline{h}(\bar{\mathbf{x}}) - \overline{g}(\bar{\mathbf{x}})|^p\right)^{1/p} \leq \left(\frac{a^p}{|S|}\sum_{\bar{\mathbf{x}} \in S} \max_{x \in \bar{\mathbf{x}}} |h(x) - g(x)|^p\right)^{1/p} \\
&\leq \left(\frac{a^p}{|S|}\sum_{\bar{\mathbf{x}} \in S}\sum_{x \in \bar{\mathbf{x}}} |h(x) - g(x)|^p\right)^{1/p} = \frac{a}{|S|^{1/p}}\left(\sum_{x \in S^{\cup}} |h(x) - g(x)|^p\right)^{1/p} \\
&= a\left(\frac{|S^{\cup}|}{|S|}\right)^{1/p}\left(\frac{1}{|S^{\cup}|}\sum_{x \in S^{\cup}} |h(x) - g(x)|^p\right)^{1/p} \\
&= ar^{1/p}\|h - g\|_{L_p(S^{\cup})} \leq ar^{1/p} \cdot \gamma.
\end{aligned}$$

It follows that $\overline{\mathcal{C}}$ is a $(ar^{1/p}\gamma)$-covering for $\overline{\mathcal{H}}$. For $p = \infty$ we have

$$\begin{aligned}
\|\overline{h} - \overline{g}\|_{L_\infty(S)} &= \max_{\bar{\mathbf{x}} \in S} |\overline{h}(\bar{\mathbf{x}}) - \overline{g}(\bar{\mathbf{x}})| \leq a \max_{\bar{\mathbf{x}} \in S}\max_{x \in \bar{\mathbf{x}}} |h(x) - g(x)| \\
&= a \max_{x \in S^{\cup}} |h(x) - g(x)| = a\|h - g\|_{L_\infty(S^{\cup})} \leq a\gamma = a \cdot r^{1/p} \cdot \gamma.
\end{aligned}$$

Thus in both cases, $\overline{\mathcal{C}}$ is a $ar^{1/p}\gamma$-covering for $\overline{\mathcal{H}}$, and its size is $\mathcal{N}(\gamma, \mathcal{H}, L_p(S^{\cup}))$. Thus

$$\mathcal{N}(ar^{1/p}\gamma, \overline{\mathcal{H}}, L_p(S^{\cup})) \leq \mathcal{N}(\gamma, \overline{\mathcal{H}}, L_p(S^{\cup})).$$

We get the statement of the lemma by substituting $\gamma$ with $\frac{\gamma}{ar^{1/p}}$. ■

As an immediate corollary, we have the following bound for covering numbers of a given sample size.

**Corollary 13** *Let $r \in \mathbb{N}$, and let $R \subseteq [r]$. Suppose the bag function $\psi : \mathbb{R}^{(R)} \to \mathbb{R}$ is $a$-Lipschitz with respect to the infinity norm for some $a > 0$. Let $\gamma > 0, p \in [1, \infty]$, and $\mathcal{H} \in \mathbb{R}^{\mathcal{X}}$. For any $m \geq 0$,*

$$\mathcal{N}_m(\gamma, \overline{\mathcal{H}}, p) \leq \mathcal{N}_{rm}(\frac{\gamma}{a \cdot r^{1/p}}, \mathcal{H}, p).$$

## 5. Margin Learning for MIL: Fat-Shattering Dimension

Large-margin classification is a popular supervised learning approach, which has received attention also as a method for MIL. For instance, MI-SVM (Andrews et al., 2002) attempts to optimize an adaptation of the soft-margin SVM objective (Cortes and Vapnik, 1995) to MIL, in which the margin of a bag is the maximal margin achieved by any of its instances. It has not been shown, however, whether minimizing the objective function of MI-SVM, or other margin formulations for MIL, allows learning with a reasonable sample size. We fill in this gap in Theorem 14 below, which bounds the $\gamma$-fat-shattering dimension (see e.g. Anthony and Bartlett 1999) of MIL. The objective of MI-SVM amounts to replacing the hypothesis class $\mathcal{H}$ of separating hyperplanes with the class of bag-hypotheses $\overline{\mathcal{H}}$ where the bag function is $\psi = \max$. Since $\max$ is the real-valued extension of OR, this objective function is natural in our MIL formulation. The distribution-free sample complexity of large-margin learning with the zero-one loss is proportional to the fat-shattering dimension (Alon et al., 1997). Thus, we provide an upper bound on the fat-shattering dimension of MIL as a function of the fat-shattering dimension of the underlying hypothesis class, and of the maximal allowed bag size. The bound holds for any Lipschitz bag-function. Let $\gamma > 0$ be the desired margin. For a hypothesis class $H$, denote its $\gamma$-fat-shattering dimension by $\mathrm{Fat}(\gamma, H)$

**Theorem 14** *Let $r \in \mathbb{N}$ and assume $R \subseteq [r]$. Let $B, a > 0$. Let $\mathcal{H} \subseteq [0, B]^{\mathcal{X}}$ be a real-valued hypothesis class and assume that the bag function $\psi : [0, B]^{(R)} \to [0, aB]$ is $a$-lipschitz with respect to the infinity norm. Then for all $\gamma \in (0, aB]$*

$$\mathrm{Fat}(\gamma, \overline{\mathcal{H}}) \leq \max\left\{33,\ 24\mathrm{Fat}(\frac{\gamma}{64a}, \mathcal{H}) \log^2\left(\frac{6 \cdot 2048 \cdot B^2 a^2}{\gamma^2} \cdot \mathrm{Fat}(\frac{\gamma}{64a}, \mathcal{H}) \cdot r\right)\right\}. \quad (6)$$

This theorem shows that for margin learning as well, the dependence of the bag size on the sample complexity is poly-logarithmic. In the proof of the theorem we use the following two results, which link the covering number of a function class with its fat-shattering dimension.

**Theorem 15 (Bartlett et al. (1997))** *Let $F$ be a set of real-valued functions and let $\gamma > 0$. For $m \geq \mathrm{Fat}(16\gamma, F)$,*

$$e^{\mathrm{Fat}(16\gamma,F)/8} \leq \mathcal{N}_m(\gamma, F, \infty).$$

The following theorem is due to Anthony and Bartlett (1999) (Theorem 12.8), following Alon et al. (1993).

**Theorem 16** *Let $F$ be a set of real-valued functions with range in $[0, B]$. Let $\gamma > 0$. For all $m \geq 1$,*

$$\mathcal{N}_m(\gamma, F, \infty) < 2\left(\frac{4B^2 m}{\gamma^2}\right)^{\mathrm{Fat}(\frac{\gamma}{4},F)\log(4eBm/\gamma)}. \quad (7)$$

Theorem 12.8 in Anthony and Bartlett (1999) deals with the case $m \geq \mathrm{Fat}(\frac{\gamma}{4}, F)$. Here we only require $m \geq 1$, since if $m \leq \mathrm{Fat}(\frac{\gamma}{4})$ then the trivial upper bound $\mathcal{N}_m(\gamma, \mathcal{H}, \infty) \leq (B/\gamma)^m \leq (B/\gamma)^{\mathrm{Fat}(\frac{\gamma}{4})}$ implies Eq. (7).

**Proof** [of Theorem 14] From Theorem 15 and Lemma 12 it follows that for $m \geq \mathrm{Fat}(16\gamma, \overline{\mathcal{H}})$,

$$\mathrm{Fat}(16\gamma, \overline{\mathcal{H}}) \leq \frac{8}{\log e} \log \mathcal{N}_m(\gamma, \overline{\mathcal{H}}, \infty) \leq 6 \log \mathcal{N}_{rm}(\gamma/a, \mathcal{H}, \infty). \quad (8)$$

By Theorem 16, for all $m \geq 1$, if $\mathrm{Fat}(\gamma/4) \geq 1$ then

$$\begin{aligned}\forall \gamma \leq \frac{B}{2e}, \quad \log \mathcal{N}_m(\gamma, \mathcal{H}, \infty) &\leq 1 + \mathrm{Fat}(\frac{\gamma}{4}, \mathcal{H}) \log(\frac{4eBm}{\gamma}) \log\left(\frac{4B^2 m}{\gamma^2}\right) \\ &\leq \mathrm{Fat}(\frac{\gamma}{4}, \mathcal{H}) \log(\frac{8eBm}{\gamma}) \log\left(\frac{4B^2 m}{\gamma^2}\right) \quad (9) \\ &\leq \mathrm{Fat}(\frac{\gamma}{4}, \mathcal{H}) \log^2(\frac{4B^2 m}{\gamma^2}). \quad (10)\end{aligned}$$

The inequality in line (9) holds since we have added 1 to the second factor, and the value of the other factors is at least 1. The last inequality follows since if $\gamma \leq \frac{B}{2e}$, we have $8eB/\gamma \leq 4B^2/\gamma^2$.

Eq. (10) also holds if $\mathrm{Fat}(\gamma/4) < 1$, since this implies $\mathrm{Fat}(\gamma/4) = 0$ and $\mathcal{N}_m(\gamma, \mathcal{H}, \infty) = 1$. Combining Eq. (8) and Eq. (10), we get that if $m \geq \mathrm{Fat}(16\gamma, \overline{\mathcal{H}})$ then

$$\forall \gamma \leq \frac{aB}{2e}, \quad \mathrm{Fat}(16\gamma, \overline{\mathcal{H}}) \leq 6\mathrm{Fat}(\frac{\gamma}{4a}, \mathcal{H}) \log^2(\frac{4B^2a^2rm}{\gamma^2}). \tag{11}$$

Set $m = \lceil \mathrm{Fat}(16\gamma, \overline{\mathcal{H}}) \rceil \leq \mathrm{Fat}(16\gamma, \overline{\mathcal{H}}) + 1$. If $\mathrm{Fat}(16\gamma, \overline{\mathcal{H}}) \geq 1$, we have that $m \geq \mathrm{Fat}(16\gamma, \overline{\mathcal{H}})$ and also $m \leq 2\mathrm{Fat}(16\gamma, \overline{\mathcal{H}})$. Thus Eq. (11) holds, and

$$\begin{aligned}\forall \gamma \leq \frac{aB}{2e}, \quad \mathrm{Fat}(16\gamma, \overline{\mathcal{H}}) &\leq 6\mathrm{Fat}(\frac{\gamma}{4a}, \mathcal{H}) \log^2(\frac{4B^2a^2}{\gamma^2} \cdot r \cdot (\mathrm{Fat}(16\gamma, \overline{\mathcal{H}}) + 1)) \\ &\leq 6\mathrm{Fat}(\frac{\gamma}{4a}, \mathcal{H}) \log^2(\frac{8B^2a^2}{\gamma^2} \cdot r \cdot \mathrm{Fat}(16\gamma, \overline{\mathcal{H}})).\end{aligned}$$

Now, it is easy to see that if $\mathrm{Fat}(16\gamma, \overline{\mathcal{H}}) < 1$, this inequality also holds. Therefore it holds in general. Substituting $\gamma$ with $\gamma/16$, we have that

$$\forall \gamma \leq \frac{8aB}{e}, \quad \mathrm{Fat}(\gamma, \overline{\mathcal{H}}) \leq 6\mathrm{Fat}(\frac{\gamma}{64a}, \mathcal{H}) \log^2(\frac{2048B^2a^2}{\gamma^2} \cdot r \cdot \mathrm{Fat}(\gamma, \overline{\mathcal{H}})). \tag{12}$$

Note that the condition on $\gamma$ holds, in particular, for all $\gamma \leq aB$.

To derive the desired Eq. (6) from Eq. (12), let $\beta = 6\mathrm{Fat}(\gamma/64a, \mathcal{H})$ and $\eta = 2048B^2a^2/\gamma^2$. Denote $F = \mathrm{Fat}(\gamma, \overline{\mathcal{H}})$. Then Eq. (12) can be restated as $F \leq \beta \log^2(\eta r F)$. It follows that $\sqrt{F}/\log(\eta r F) \leq \sqrt{\beta}$, Thus

$$\frac{\sqrt{F}}{\log(\eta r F)} \log\left(\frac{\sqrt{\eta r F}}{\log(\eta r F)}\right) \leq \sqrt{\beta} \log(\sqrt{\beta \eta r}).$$

Therefore

$$\frac{\sqrt{F}}{\log(\eta r F)} (\log(\eta r F)/2 - \log(\log(\eta r F))) \leq \sqrt{\beta} \log(\beta \eta r)/2,$$

hence

$$(1 - \frac{2\log(\log(\eta r F))}{\log(\eta r F)})\sqrt{F} \leq \sqrt{\beta} \log(\beta \eta r).$$

Now, it is easy to verify that $\log(\log(x))/\log(x) \leq \frac{1}{4}$ for all $x \geq 33 \cdot 2048$. Assume $F \geq 33$ and $\gamma \leq aB$. Then

$$\eta r F = 2048B^2a^2rF/\gamma^2 \geq 2048F \geq 33 \cdot 2048.$$

Therefore $\log(\log(\eta r F))/\log(\eta r F) \leq \frac{1}{4}$, which implies $\frac{1}{2}\sqrt{F} \leq \sqrt{\beta} \log(\beta \eta r)$. Thus $F \leq 4\beta \log^2(\beta \eta r)$. Substituting the parameters with their values, we get the desired bound, stated in Eq. (6). ■

## 6. Sample Complexity by Average Bag Size: Rademacher Complexity

The upper bounds we have shown so far provide distribution-free sample complexity bounds, which depend only on the maximal possible bag size. In this section we show that even if the bag size is unbounded, we can still have a sample complexity guarantee, if the *average* bag size for the input distribution is bounded. For this analysis we use the notion of Rademacher complexity (Bartlett and Mendelson, 2002). Let $A$ be some domain. The empirical Rademacher complexity of a class of functions $\mathcal{F} \subseteq \mathbb{R}^{A\times\{-1,+1\}}$ with respect to a sample $S = \{(x_i, y_i)\}_{i\in[m]} \subseteq A\times\{-1, +1\}$ is

$$\mathcal{R}(\mathcal{F}, S) \triangleq \frac{1}{m}\mathbb{E}_\sigma[|\sup_{f\in\mathcal{F}} \sum_{i\in[m]} \sigma_i f(x_i, y_i)|],$$

where $\sigma = (\sigma_1, \ldots, \sigma_m)$ are $m$ independent uniform $\{\pm1\}$-valued variables. The average Rademacher complexity of $\mathcal{F}$ with respect to a distribution $D$ over $A \times \{-1, +1\}$ and a sample size $m$ is

$$\mathcal{R}_m(\mathcal{F}, D) \triangleq \mathbb{E}_{S\sim D^m}[\mathcal{R}(\mathcal{F}, S)].$$

The worst-case Rademacher complexity over samples of size $m$ is

$$\mathcal{R}_m^{\sup}(\mathcal{F}) = \sup_{S\subseteq A^m} \mathcal{R}(\mathcal{F}, S).$$

This quantity can be tied to the fat-shattering dimension via the following result:

**Theorem 17 (See e.g. Mendelson (2002), Theorem 4.11)** *Let $m \geq 1$ and $\gamma \geq 0$. If $\mathcal{R}_m^{\sup}(\mathcal{F}) \leq \gamma$ then the $\gamma$-fat-shattering dimension of $\mathcal{F}$ is at most $m$.*

Let $I \subseteq \mathbb{R}$. Assume a hypothesis class $H \subseteq I^A$ and a loss function $\ell : \{-1, +1\} \times I \to \mathbb{R}$. For a hypothesis $h \in H$, we denote by $h_\ell$ the function defined by $h_\ell(x, y) = \ell(y, h(x))$. Given $H$ and $\ell$, we define the function class $H_\ell \triangleq \{h_\ell \mid h \in H\} \subseteq \mathbb{R}^{A\times\{-1,+1\}}$.

Rademacher complexities can be used to derive sample complexity bounds (Bartlett and Mendelson, 2002): Assume the range of the loss function is $[0, 1]$. For any $\delta \in (0, 1)$, with probability of $1 - \delta$ over the draw of samples $S \subseteq A \times \{-1, +1\}$ of size $m$ drawn from $D$, every $h \in H$ satisfies

$$\ell(h, D) \leq \ell(h, S) + 2\mathcal{R}_m(H_\ell, D) + \sqrt{\frac{8\ln(2/\delta)}{m}}. \tag{13}$$

Thus, an upper bound on the Rademacher complexity implies an upper bound on the average loss of a classifier learned from a random sample.

### 6.1 Binary MIL

Our first result complements the distribution-free sample complexity bounds that were provided for binary MIL in Section 3. The average (or expected) bag size under a distribution $D$ over $\mathcal{X}^{(R)} \times \{-1, +1\}$ is $\mathbb{E}_{(\bar{\mathbf{X}},Y)\sim D}[|\bar{\mathbf{X}}|]$. Our sample complexity bound for binary MIL depends on the average bag size and the VC dimension of the instance hypothesis class. Recall that the zero-one loss is defined by $\ell_{0/1}(y, \hat{y}) = \mathbb{1}[y \neq \hat{y}]$. For a sample of labeled examples $S = \{(x_i, y_i)\}_{i\in[m]}$, we use $S_X$ to denote the examples of $S$, that is $S_X = \{x_i\}_{i\in[m]}$.

**Theorem 18** *Let $\mathcal{H} \subseteq \{-1,+1\}^{\mathcal{X}}$ be a binary hypothesis class with VC-dimension $d$. Let $R \subseteq \mathbb{N}$ and assume a bag function $\psi : \{-1,+1\}^{(R)} \to \{-1,+1\}$. Let $r$ be the average bag size under distribution $D$ over labeled bags. Then*

$$\mathcal{R}(\overline{\mathcal{H}}_{\ell_{0/1}}, D) \leq 17\sqrt{\frac{d\ln(4er)}{m}}.$$

**Proof** Let $S$ be a labeled bag-sample of size $m$. Dudley's entropy integral (Dudley, 1967) states that

$$\begin{aligned}\mathcal{R}(\overline{\mathcal{H}}_{\ell_{0/1}}, S) &\leq \frac{12}{\sqrt{m}}\int_0^\infty \sqrt{\ln \mathcal{N}(\gamma, \overline{\mathcal{H}}_{\ell_{0/1}}, L_2(S))}\, d\gamma \qquad (14)\\ &= \frac{12}{\sqrt{m}}\int_0^1 \sqrt{\ln \mathcal{N}(\gamma, \overline{\mathcal{H}}_{\ell_{0/1}}, L_2(S))}\, d\gamma.\end{aligned}$$

The second equality holds since for any $\gamma > 1$, $\mathcal{N}(\gamma, \overline{\mathcal{H}}_{\ell_{0/1}}, L_2(S)) = 1$, thus the expression in the integral is zero.

If $\mathcal{C}$ is a $\gamma$-cover for $\overline{\mathcal{H}}$ with respect to the norm $L_2(S_X)$, then $\mathcal{C}_{\ell_{0/1}}$ is a $\gamma/2$-cover for $\overline{\mathcal{H}}_{\ell_{0/1}}$ with respect to the norm $L_2(S)$. This can be seen as follows: Let $h_{\ell_{0/1}} \in \overline{\mathcal{H}}_{\ell_{0/1}}$ for some $h \in \overline{\mathcal{H}}$. Let $f \in \mathcal{C}$ such that $\|f - h\|_{L_2(S_X)} \leq \gamma$. We have

$$\begin{aligned}\|f_{\ell_{0/1}} - h_{\ell_{0/1}}\|_{L_2(S)} &= \left(\frac{1}{m}\sum_{(x,y)\in S} |f_{\ell_{0/1}}(x,y) - h_{\ell_{0/1}}(x,y)|^2\right)^{1/2}\\ &= \left(\frac{1}{m}\sum_{(x,y)\in S} |\ell_{0/1}(y, f(x)) - \ell_{0/1}(y, h(x))|^2\right)^{1/2}\\ &= \left(\frac{1}{m}\sum_{x\in S_X} (\frac{1}{2}|f(x) - h(x)|)^2\right)^{1/2} = \frac{1}{2}\|f - h\|_{L_2(S_X)} \leq \gamma/2.\end{aligned}$$

Therefore $\mathcal{C}_{\ell_{0/1}}$ is a $\gamma/2$-cover for $L_2(S)$. It follows that we can bound the $\gamma$-covering number of $\overline{\mathcal{H}}_{\ell_{0/1}}$ by:

$$\mathcal{N}(\gamma, \overline{\mathcal{H}}_{\ell_{0/1}}, L_2(S)) \leq \mathcal{N}(2\gamma, \overline{\mathcal{H}}, L_2(S_X)). \qquad (15)$$

Let $r(S)$ be the average bag size in the sample $S$, that is $r(S) = |S^\cup|/|S|$. By Lemma 12,

$$\mathcal{N}(\gamma, \overline{\mathcal{H}}, L_2(S_X)) \leq \mathcal{N}(\gamma/\sqrt{r(S)}, \mathcal{H}, L_2(S_X^\cup)). \qquad (16)$$

From Eq. (14), Eq. (15) and Eq. (16) we conclude that

$$\mathcal{R}(\overline{\mathcal{H}}_{\ell_{0/1}}, S) \leq \frac{12}{\sqrt{m}}\int_0^1 \sqrt{\ln \mathcal{N}(2\gamma/\sqrt{r(S)}, \mathcal{H}, L_2(S_X^\cup))}\, d\gamma.$$

By Dudley (1978), for any $\mathcal{H}$ with VC-dimension $d$, and any $\gamma > 0$,

$$\ln \mathcal{N}(\gamma, \mathcal{H}, L_2(S_X^\cup)) \leq 2d\ln\left(\frac{4e}{\gamma^2}\right).$$

Therefore

$$\begin{aligned}\mathcal{R}(\overline{\mathcal{H}}_{\ell_{0/1}}, S) &\leq \frac{12}{\sqrt{m}} \int_0^1 \sqrt{2d \ln\left(\frac{er(S)}{\gamma^2}\right)}\, d\gamma \\ &\leq 17\sqrt{\frac{d}{m}} \left(\int_0^1 \sqrt{\ln(er(S))}\, d\gamma + \int_0^1 \sqrt{\ln(1/\gamma^2)}\, d\gamma\right) \\ &= 17\sqrt{\frac{d(\ln(er(S)) + \sqrt{\pi/2})}{m}} \leq 17\sqrt{\frac{d\ln(4er(S))}{m}}.\end{aligned}$$

The function $\sqrt{\ln(x)}$ is concave for $x \geq 1$. Therefore we may take the expectation of both sides of this inequality and apply Jensen's inequality, to get

$$\begin{aligned}\mathcal{R}_m(\overline{\mathcal{H}}_{\ell_{0/1}}, D) = \mathbb{E}_{S\sim D^m}[\mathcal{R}(\overline{\mathcal{H}}_{\ell_{0/1}}, S)] &\leq \mathbb{E}_{S\sim D^m}\left[17\sqrt{\frac{d\ln(4er(S))}{m}}\right] \\ &\leq 17\sqrt{\frac{d\ln(4e\cdot\mathbb{E}_{S\sim D^m}[r(S)])}{m}} = 17\sqrt{\frac{d\ln(4er)}{m}}.\end{aligned}$$

■

We conclude that even when the bag size is not bounded, the sample complexity of binary MIL with a specific distribution depends only logarithmically on the average bag size in this distribution, and linearly on the VC-dimension of the underlying instance hypothesis class.

### 6.2 Real-Valued Hypothesis Classes

In our second result we wish to bound the sample complexity of MIL when using other loss functions that accept real valued predictions. This bound will depend on the average bag size, and on the Rademacher complexity of the instance hypothesis class.

We consider the case where both the bag function and the loss function are Lipschitz. For the bag function, recall that all extensions of monotone Boolean functions are Lipschitz with respect to the infinity norm. For the loss function $\ell : \{-1,+1\}\times\mathbb{R}\to\mathbb{R}$, we require that it is Lipschitz in its second argument, i.e. that there is a constant $a > 0$ such that for all $y \in \{-1,+1\}$ and $y_1, y_2 \in \mathbb{R}$, $|\ell(y,y_1) - \ell(y,y_2)| \leq a|y_1 - y_2|$. This property is satisfied by many popular losses. For instance, consider the hinge-loss, which is the loss minimized by soft-margin SVM. It is defined as $\ell_{hl}(y,\hat{y}) = [1 - y\hat{y}]_+$, and is 1-Lipschitz in its second argument.

The following lemma provides a bound on the empirical Rademacher complexity of MIL, as a function of the average bag size in the sample and of the behavior of the worst-case Rademacher complexity over instances. We will subsequently use this bound to bound the average Rademacher complexity of MIL with respect to a distribution. We consider losses with the range $[0,1]$. To avoid degenerate cases, we consider only losses such that there exists at least one labeled bag $(\bar{\mathbf{x}}, y) \subseteq \mathcal{X}^{(R)} \times \{-1,+1\}$ and hypotheses $h, g \in \mathcal{H}$ such that $h_\ell(\bar{\mathbf{x}}, y) = 0$ and $g_\ell(\bar{\mathbf{x}}, y) = 1$. We say that such a loss has a *full range*.

**Lemma 19** *Let $\mathcal{H} \subseteq [0,B]^{\mathcal{X}}$ be a hypothesis class. Let $R \subseteq \mathbb{N}$, and let the bag function $\psi : \mathbb{R}^{(R)} \to \mathbb{R}$ be $a_1$-Lipschitz with respect to the infinity norm. Assume a loss function $\ell : \{-1,+1\} \times$* $\mathbb{R} \to [0,1]$, *which is $a_2$-Lipschitz in its second argument. Further assume that $\ell$ has a full range. Suppose there is a continuous decreasing function $f : (0,1] \to \mathbb{R}$ such that*

$$\forall \gamma \in (0,1], \quad f(\gamma) \in \mathbb{N} \implies \mathcal{R}^{\sup}_{f(\gamma)}(\mathcal{H}) \leq \gamma.$$

*Let $S$ be a labeled bag-sample of size $m$, with an average bag size $r$. Then for all $\epsilon \in (0,1]$,*

$$\mathcal{R}(\overline{\mathcal{H}}_\ell, S) \leq 4\epsilon + \frac{10}{\sqrt{m}} \log\left(\frac{4ea_1^2a_2^2B^2rm}{\epsilon^2}\right)\left(1 + \int_\epsilon^1 \sqrt{f(\frac{\gamma}{4a_1a_2})}\, d\gamma\right).$$

**Proof** A refinement of Dudley's entropy integral (Srebro et al., 2010, Lemma A.3) states that for all $\epsilon \in (0,1]$, for all real function classes $\mathcal{F}$ with range $[0,1]$ and for all sets $S$,

$$\mathcal{R}(\mathcal{F}, S) \leq 4\epsilon + \frac{10}{\sqrt{m}} \int_\epsilon^1 \sqrt{\ln \mathcal{N}(\gamma, \mathcal{F}, L_2(S))}\, d\gamma. \tag{17}$$

Since the range of $\ell$ is $[0,1]$, this holds for $\mathcal{F} = \overline{\mathcal{H}}_\ell$. In addition, for any set $S$, the $L_2(S)$ norm is bounded from above by the $L_\infty(S)$ norm. Therefore $\mathcal{N}(\gamma, \mathcal{F}, L_2(S)) \leq \mathcal{N}(\gamma, \mathcal{F}, L_\infty(S))$. Thus, by Eq. (17) we have

$$\mathcal{R}(\overline{\mathcal{H}}_\ell, S) \leq 4\epsilon + \frac{10}{\sqrt{m}} \int_\epsilon^1 \sqrt{\ln \mathcal{N}(\gamma, \overline{\mathcal{H}}_\ell, L_\infty(S))}\, d\gamma. \tag{18}$$

Now, let $h, g \in \mathcal{H}$ and consider $\overline{h}_\ell, \overline{g}_\ell \in \overline{\mathcal{H}}_\ell$. Since $\ell$ is $a_2$-Lipschitz, we have

$$\begin{aligned}
\|\overline{h}_\ell - \overline{g}_\ell\|_{L_\infty(S)} &= \max_{i\in[m]} |\overline{h}_\ell(\bar{\mathbf{x}}_i, y_i) - \overline{g}_\ell(\bar{\mathbf{x}}_i, y_i)| = \max_{i\in[m]} |\ell(y_i, \overline{h}(\bar{\mathbf{x}}_i)) - \ell(y_i, \overline{g}(\bar{\mathbf{x}}_i))| \\
&\leq a_2 \max_{i\in[m]} |\overline{h}(\bar{\mathbf{x}}_i) - \overline{g}(\bar{\mathbf{x}}_i)| = a_2\|\overline{h} - \overline{g}\|_{L_\infty(S_X)}.
\end{aligned}$$

It follows that if $\mathcal{C} \subseteq \overline{\mathcal{H}}$ is a $\gamma/a_2$-cover for $\overline{\mathcal{H}}$ then $\mathcal{C}_\ell \subseteq \overline{\mathcal{H}}_\ell$ is a $\gamma$-cover for $\overline{\mathcal{H}}_\ell$. Therefore $\mathcal{N}(\gamma, \overline{\mathcal{H}}_\ell, L_\infty(S)) \leq \mathcal{N}(\gamma/a_2, \overline{\mathcal{H}}, L_\infty(S_X))$. By Lemma 12,

$$\mathcal{N}(\gamma/a_2, \overline{\mathcal{H}}, L_\infty(S_X)) \leq \mathcal{N}(\gamma/a_1a_2, \mathcal{H}, L_\infty(S_X^\cup)) \leq \mathcal{N}_{rm}(\gamma/a_1a_2, \mathcal{H}, \infty).$$

Combining this with Eq. (18) it follows that

$$\mathcal{R}(\overline{\mathcal{H}}_\ell, S) \leq 4\epsilon + \frac{10}{\sqrt{m}} \int_\epsilon^1 \sqrt{\mathcal{N}_{rm}(\gamma/a_1a_2, \mathcal{H}, \infty)}\, d\gamma. \tag{19}$$

Now, let $\gamma \in (0,1]$, and let $\gamma_\circ = \sup\{\gamma_\circ \leq \gamma \mid f(\gamma_\circ) \in \mathbb{N}\}$. Since $\mathcal{R}^{\sup}_{f(\gamma_\circ)}(\mathcal{H}) \leq \gamma_\circ$, by Theorem 17 the $\gamma_\circ$-fat-shattering dimension of $\mathcal{H}$ is at most $f(\gamma_\circ)$. It follows that

$$\mathrm{Fat}(\gamma, \mathcal{H}) \leq \mathrm{Fat}(\gamma_\circ, \mathcal{H}) \leq f(\gamma_\circ) \leq 1 + f(\gamma).$$

The last inequality follows from the definition of $\gamma_\circ$, since $f$ is continuous and decreasing. Therefore, by Theorem 16,

$$\begin{aligned}
\forall \gamma \leq B, \quad \log \mathcal{N}_m(\gamma, \mathcal{H}, \infty) &\leq 1 + (f(\frac{\gamma}{4}) + 1)\log(\frac{4eBm}{\gamma})\log\left(\frac{4B^2m}{\gamma^2}\right) \\
&\leq (f(\frac{\gamma}{4}) + 1)\log(\frac{4eBm}{\gamma})\log\left(\frac{4eB^2m}{\gamma^2}\right) &(20)\\
&\leq (f(\frac{\gamma}{4}) + 1)\log^2(\frac{4eB^2m}{\gamma^2}). &(21)
\end{aligned}$$

The inequality in line (20) holds since we have added $\log(e) \geq 1$ to the third factor, and the value of the other factors is at least 1. The last inequality follows since $\gamma \leq B$.

We now show that the assumption $\gamma \leq B$ does not restrict us: By the assumptions on $\ell$, there are $h, g \in \mathcal{H}$ and a labeled bag $(\bar{\mathbf{x}}, y)$ such that $\overline{h}_\ell(\bar{\mathbf{x}}, y) = 1$ and $\overline{g}_\ell(\bar{\mathbf{x}}, y) = 0$. Let $n = |\bar{\mathbf{x}}|$. By the Lipschitz assumptions we have

$$\begin{aligned}
1 &= |\overline{h}_\ell(\bar{\mathbf{x}}, y) - \overline{g}_\ell(\bar{\mathbf{x}}, y)| = |\ell(y, \overline{h}(\bar{\mathbf{x}})) - \ell(y, \overline{g}(\bar{\mathbf{x}}))| \leq a_2|\overline{h}(\bar{\mathbf{x}}) - \overline{g}(\bar{\mathbf{x}})| \\
&= a_2|\psi(h(x[1]), \ldots, h(x[n])) - \psi(g(x[1]), \ldots, g(x[n]))| \leq a_2 a_1 \max_{j \in [n]} |h(x[j]) - g(x[j])| \leq a_1 a_2 B.
\end{aligned}$$

Thus $1 \leq a_1 a_2 B$. It follows that for all $\gamma \in (0, 1]$, $\gamma / a_1 a_2 \leq B$. Thus Eq. (21) can be combined with Eq. (19) to get that for all $\epsilon \in (0, 1]$,

$$\begin{aligned}
\mathcal{R}(\overline{\mathcal{H}}_\ell, S) &\leq 4\epsilon + \frac{10}{\sqrt{m}} \int_\epsilon^1 \sqrt{\left(f(\frac{\gamma}{4a_1 a_2}) + 1\right) \log^2\left(\frac{4ea_1^2 a_2^2 B^2 rm}{\gamma^2}\right)}\, d\gamma \\
&\leq 4\epsilon + \frac{10}{\sqrt{m}} \log\left(\frac{4ea_1^2 a_2^2 B^2 rm}{\epsilon^2}\right) \int_\epsilon^1 \sqrt{f(\frac{\gamma}{4a_1 a_2}) + 1}\, d\gamma \\
&\leq 4\epsilon + \frac{10}{\sqrt{m}} \log\left(\frac{4ea_1^2 a_2^2 B^2 rm}{\epsilon^2}\right) \left(1 + \int_\epsilon^1 \sqrt{f(\frac{\gamma}{4a_1 a_2})}\, d\gamma\right).
\end{aligned}$$

The last inequality follows from the fact that $\sqrt{a+b} \leq \sqrt{a} + \sqrt{b}$ for non-negative $a$ and $b$, and from $\int_\epsilon^1 1 \leq 1$. ■

Based on Lemma 19, we will now bound the average Rademacher complexity of MIL, as a function of the worst-case Rademacher complexity over instances, and the expected bag size. Since the number of instances in a bag sample of a certain size is not fixed, but depends on the bag sizes in the specific sample, we will need to consider the behavior of $\mathcal{R}_m^{\sup}(\mathcal{H})$ for different values of $m$. For many learnable function classes, the Rademacher complexity is proportional to $\frac{1}{\sqrt{m}}$, or to $\frac{ln^\beta(m)}{\sqrt{m}}$ for some non-negative $\beta$. The following theorem bounds the average Rademacher complexity of MIL in all these cases. The resulting bound indicates that here too there is a poly-logarithmic dependence of the sample complexity on the average bag size. Following the proof we show an application of the bound to a specific function class.

**Theorem 20** *Let $\mathcal{H} \subseteq [0, B]^\mathcal{X}$ be a hypothesis class. Let $R \subseteq \mathbb{N}$, and let the bag function $\psi : \mathbb{R}^{(R)} \to \mathbb{R}$ be $a_1$-Lipschitz with respect to the infinity norm. Assume a loss function $\ell : \{-1, +1\} \times \mathbb{R} \to [0, 1]$, which is $a_2$-Lipschitz in its second argument. Further assume that $\ell$ has a full range. Suppose that there are $C, \beta, K \geq 0$ such that for all $m \geq K$,*

$$\mathcal{R}_m^{\sup}(\mathcal{H}) \leq \frac{C \ln^\beta(m)}{\sqrt{m}}.$$

*Then there exists a number $N \geq 0$ that depends only on $C, \beta$ and $K$ such that for any distribution $D$ with average bag size $r$, and for all $m \geq 1$,*

$$\mathcal{R}_m(\overline{\mathcal{H}}_\ell, D) \leq \frac{4 + 10\log(4ea_1^2 a_2^2 B^2 rm^2)\left(N + \frac{a_1 a_2}{\beta+1} C \ln^{\beta+1}(16a_1^2 a_2^2 m)\right)}{\sqrt{m}}.$$

**Proof** Let $S$ be a labeled bag sample of size $m$, and let $\tilde{r}$ be its average bag size. Denote $T(x) = C\ln^{\beta}(x)$, and define $f(\gamma) = \frac{4T^2(1/\gamma^2)}{\gamma^2}$. We will show that $\mathcal{R}^{\sup}_{f(\gamma)} \leq \gamma$, thus allowing the use of Lemma 19. We have $\mathcal{R}_m \leq T(m)/\sqrt{m}$, thus it suffices to show that $T(f(\gamma))/\sqrt{f(\gamma)} \leq \gamma$.

Let $z(\gamma) = \sqrt{f(\gamma)}/T(f(\gamma))$. We will now show that $z(\gamma)T(z^2(\gamma)) \geq \frac{1}{\gamma}T(1/\gamma^2)$. Since the function $xT(x^2) = Cx\ln^{\beta}(x^2)$ is monotonic increasing for $x \geq 1$, we will conclude that $z(\gamma) \geq 1/\gamma$ for all $\gamma \leq 1$.

It is easy to see that for all values of $\beta, C \geq 0$, there is a number $n \geq 0$ such that for all $x \geq n$,

$$C^2\ln^{2\beta}(x) \leq x^{1-2^{-1/\beta}}.$$

For such $x$ we have

$$\begin{aligned} T(x/T^2(x)) &= C\ln^{\beta}\left(\frac{x}{C^2\ln^{2\beta}(x)}\right) = C(\ln(x) - \ln(C^2\ln^{2\beta}(x)))^{\beta} \\ &\geq C(\ln(x) - (1-2^{-1/\beta})\ln(x)))^{\beta} = C\ln^{\beta}(x)/2 = T(x)/2. \end{aligned} \tag{22}$$

Let $\gamma_\circ \in (0,1)$ such that $f(\gamma_\circ) = k = \max\{n, K\}$. Since $f(\gamma)$ is monotonic decreasing with $\gamma$, for all $\gamma \leq \gamma_\circ$, $f(\gamma) \geq k$. Therefore, for $\gamma \leq \gamma_\circ$,

$$z(\gamma)T(z^2(\gamma)) = \frac{\sqrt{f(\gamma)}}{T(f(\gamma))}T\left(\frac{f(\gamma)}{T^2(f(\gamma))}\right) \geq \frac{1}{2}\frac{\sqrt{f(\gamma)}}{T(f(\gamma))}T(f(\gamma)) = \frac{1}{2}\sqrt{f(\gamma)} = T(1/\gamma^2)/\gamma.$$

The middle inequality follows from Eq. (22), and the last equality follows from the definition of $f(\gamma)$. We conclude that $z(\gamma) \geq \frac{1}{\gamma}$. Therefore, for all $\gamma \leq \gamma_\circ$,

$$\mathcal{R}^{\sup}_{f(\gamma)}(\mathcal{H}) \leq \frac{T(f(\gamma))}{\sqrt{f(\gamma)}} = 1/z(\gamma) \leq \gamma.$$

Define $\tilde{f}$ as follows:

$$\tilde{f}(\gamma) = \begin{cases} f(\gamma) & \gamma \leq \gamma_\circ \\ k & \gamma > \gamma_\circ. \end{cases}$$

For $\gamma \leq \gamma_\circ$, clearly $\mathcal{R}^{\sup}_{\tilde{f}(\gamma)}(\mathcal{H}) \leq \gamma$, and for $\gamma > \gamma_\circ$,

$$\mathcal{R}^{\sup}_{\tilde{f}(\gamma)}(\mathcal{H}) = \mathcal{R}^{\sup}_{k}(\mathcal{H}) = \mathcal{R}^{\sup}_{f(\gamma_\circ)}(\mathcal{H}) \leq \gamma_\circ \leq \gamma.$$

Therefore for all $\gamma \in (0,1]$, $\mathcal{R}^{\sup}_{\tilde{f}(\gamma)}(\mathcal{H}) \leq \gamma$. By Lemma 19, for all $\epsilon \in (0,1]$,

$$\begin{aligned} \mathcal{R}(\overline{\mathcal{H}}_\ell, S) &\leq 4\epsilon + \frac{10}{\sqrt{m}}\log\left(\frac{4ea_1^2a_2^2B^2\tilde{r}m}{\epsilon^2}\right)\left(1 + \int_{\epsilon}^{1}\sqrt{\tilde{f}\left(\frac{\gamma}{4a_1a_2}\right)}\,d\gamma\right) \\ &= 4\epsilon + \frac{10}{\sqrt{m}}\log\left(\frac{4ea_1^2a_2^2B^2\tilde{r}m}{\epsilon^2}\right)\left(1 + \int_{4a_1a_2\gamma_\circ}^{1}\sqrt{k}\,d\gamma + \int_{\epsilon}^{4a_1a_2\gamma_\circ}\sqrt{f\left(\frac{\gamma}{4a_1a_2}\right)}\,d\gamma\right) \\ &\leq 4\epsilon + \frac{10}{\sqrt{m}}\log\left(\frac{4ea_1^2a_2^2B^2\tilde{r}m}{\epsilon^2}\right)\left(1 + \sqrt{k} + \int_{\epsilon}^{4a_1a_2\gamma_\circ}\sqrt{f\left(\frac{\gamma}{4a_1a_2}\right)}\,d\gamma\right). \end{aligned} \tag{23}$$

Denote $N = 1 + \sqrt{k}$. Now, if $\beta > 0$ we have

$$\int_{\epsilon}^{4a_1a_2\gamma_\circ} \sqrt{f(\frac{\gamma}{4a_1a_2})}\, d\gamma \le \int_{\epsilon}^{4a_1a_2\gamma_\circ} \sqrt{f(\frac{\gamma}{4a_1a_2})}\, d\gamma = 2a_1a_2 \int_{\epsilon}^{4a_1a_2\gamma_\circ} \frac{T(16a_1^2a_2^2/\gamma^2)}{\gamma}\, d\gamma$$
$$= 2a_1a_2C \int_{\epsilon}^{4a_1a_2\gamma_\circ} \frac{\ln^{\beta}(16a_1^2a_2^2/\gamma^2)}{\gamma}\, d\gamma = 2a_1a_2C \left[-\ln^{\beta+1}(\frac{16a_1^2a_2^2}{\gamma^2})/(2(\beta+1))\right]_{\epsilon}^{4a_1a_2\gamma_\circ}$$
$$= \frac{a_1a_2C}{\beta+1} \ln^{\beta+1}(\frac{16a_1^2a_2^2\gamma_\circ^2}{\epsilon^2}) \le \frac{a_1a_2C}{\beta+1} \left(\ln^{\beta+1}(\frac{16a_1^2a_2^2}{\epsilon^2})\right).$$

The same inequality holds also for $\beta = 0$, since in that case

$$\int_{\epsilon}^{4a_1a_2\gamma_\circ} \sqrt{f(\frac{\gamma}{4a_1a_2})}\, d\gamma = 2a_1a_2 \int_{\epsilon}^{4a_1a_2\gamma_\circ} \frac{T(16a_1^2a_2^2/\gamma^2)}{\gamma}\, d\gamma$$
$$= 2a_1a_2C \int_{\epsilon}^{4a_1a_2\gamma_\circ} \frac{1}{\gamma}\, d\gamma = 2a_1a_2C\,[\ln(\gamma)]_{\epsilon}^{4a_1a_2\gamma_\circ} = 2a_1a_2C \ln(\frac{4a_1a_2\gamma_\circ}{\epsilon})$$
$$\le 2a_1a_2C \ln(\frac{4a_1a_2}{\epsilon}) = \frac{a_1a_2C}{\beta+1} \left(\ln^{\beta+1}(\frac{16a_1^2a_2^2}{\epsilon^2})\right).$$

Therefore we can further bound Eq. (23) to get

$$\mathcal{R}(\overline{\mathcal{H}}_\ell, S) \le 4\epsilon + \frac{10}{\sqrt{m}} \log\left(\frac{4ea_1^2a_2^2B^2\tilde{r}m}{\epsilon^2}\right) \left(N + \frac{a_1a_2C}{\beta+1} \ln^{\beta+1}(\frac{16a_1^2a_2^2}{\epsilon^2})\right).$$

Setting $\epsilon = 1/\sqrt{m}$ we get

$$\mathcal{R}(\overline{\mathcal{H}}_\ell, S) \le \frac{4 + 10\log(4ea_1^2a_2^2B^2\tilde{r}m^2)\left(N + \frac{a_1a_2C}{\beta+1}\ln^{\beta+1}(16a_1^2a_2^2m)\right)}{\sqrt{m}}.$$

Now, for a given sample $S$ denote its average bag size by $\tilde{r}(S)$. We have

$$\mathcal{R}_m(\overline{\mathcal{H}}_\ell, D) = \mathbb{E}_{S\sim D^m}[\mathcal{R}(\overline{\mathcal{H}}_\ell, S)]$$
$$\le \mathbb{E}\left[\frac{4 + 10\log(4ea_1^2a_2^2B^2\tilde{r}(S)m^2)\left(N + \frac{a_1a_2C}{\beta+1}\ln^{\beta+1}(16a_1^2a_2^2m)\right)}{\sqrt{m}}\right]$$
$$\le \frac{4 + 10\log(4ea_1^2a_2^2B^2rm^2)\left(N + \frac{a_1a_2C}{\beta+1}\ln^{\beta+1}(16a_1^2a_2^2m)\right)}{\sqrt{m}}.$$

In the last inequality we used Jensen's inequality and the fact that $\mathbb{E}_{S\sim D^m}[\tilde{r}(S)] = r$. This is the desired bound, hence the theorem is proven. ■

To demonstrate the implications of this theorem, consider the case of MIL with soft-margin kernel SVM. Kernel SVM can operate in a general Hilbert space, which we denote by $\mathcal{T}$. The domain of instances is $\mathcal{X} = \{x \in \mathcal{T} \mid \|x\| \le 1\}$, and the function class is the class of linear separators with a bounded norm $\mathcal{W}(C) = \{h_w \mid w \in \mathcal{T}, \|w\| \le C\}$, for some $C > 0$, where $h_w = \langle x, w\rangle$.

The loss is the hinge-loss $\ell_{hl}$ defined above, which is 1-Lipschitz in the second argument. We have (Bartlett and Mendelson, 2002)

$$\mathcal{R}^{\sup}_m(\mathcal{W}(C)_{\ell_{hl}}) \leq \frac{C}{\sqrt{m}} = \frac{C \ln^0(m)}{\sqrt{m}}.$$

Thus we can apply Theorem 20 with $\beta = 0$. Note that $\mathcal{W}(C) \subseteq [-C, C]^{\mathcal{X}}$, thus we can apply the theorem with $B = 2C$ by simply shifting the output of each $h_w$ by $C$ and adjusting the loss function accordingly. By Theorem 20 there exists a number $N$ such that for any 1-Lipschitz bag-function $\psi$ (such as $\max$) and for any distribution $D$ over labeled bags with an average bag size of $r$, we have

$$\mathcal{R}_m(\overline{\mathcal{H}}_\ell, D) \leq \frac{4 + 10 \log(16eC^2rm^2)\,(N + C\ln(16m))}{\sqrt{m}}.$$

We can use this result and apply Eq. (13) to get an upper bound on the loss of MIL with soft-margin SVM.

## 7. PAC-Learning for MIL

In the previous sections we addressed the sample complexity of generalized MIL, showing that it grows only logarithmically with the bag size. We now turn to consider the computational aspect of MIL, and specifically the relationship between computational feasibility of MIL and computational feasibility of the learning problem for the underlying instance hypothesis.

We consider real-valued hypothesis classes $\mathcal{H} \in [-1, +1]^{\mathcal{X}}$, and provide a MIL algorithm which uses a learning algorithm that operates on single instances as an oracle. We show that if the oracle can minimize error with respect to $\mathcal{H}$, and the bag-function satisfies certain boundedness conditions, then the MIL algorithm is guaranteed to PAC-learn $\overline{\mathcal{H}}$. In particular, the guarantees hold if the bag-function is Boolean OR or $\max$, as in classical MIL and its extension to real-valued hypotheses.

Given an algorithm $\mathcal{A}$ that learns $\mathcal{H}$ from single instances, we provide an algorithm called `MILearn` that uses $\mathcal{A}$ to implement a *weak learner* for bags with respect to $\overline{\mathcal{H}}$. That is, for any weighted sample of bags, `MILearn` returns a hypothesis from $\overline{\mathcal{H}}$ that has some success in labeling the bag-sample correctly. This will allow the use of `MILearn` as the building block in a Boosting algorithm (Freund and Schapire, 1997), which will find a linear combination of hypotheses from $\overline{\mathcal{H}}$ that classifies unseen bags with high accuracy. Furthermore, if $\mathcal{A}$ is efficient then the resulting Boosting algorithm is also efficient, with a polynomial dependence on the maximal bag size.

We open with background on Boosting in Section 7.1. We then describe the weak learner in and analyze its properties in Section 7.2. In Section 7.3 we provide guarantees on a Boosting algorithm that uses our weak leaner, and conclude that the computational complexity of PAC-learning for MIL can be bounded by the computational complexity of agnostic PAC-learning for single instances.

### 7.1 Background: Boosting with Margin Guarantees

In this section we give some background on Boosting algorithms, which we will use to derive an efficient learning algorithm for MIL. Boosting methods (Freund and Schapire, 1997) are techniques that allow enhancing the power of a *weak learner*—a learning algorithm that achieves error slightly better than chance—to derive a classification rule that has low error on an input sample. The idea is to iteratively execute the weak learner on weighted versions of the input sample, and then to return a linear combination of the classifiers that were emitted by the weak learner in each round.

Let $A$ be a domain of objects to classify, and let $H : [-1,+1]^A$ be the hypothesis class used by the weak learner. A Boosting algorithm receives as input a labeled sample $S = \{(x_i, y_i)\}_{i=1}^m \subseteq A \times \{-1,+1\}$, and iteratively feeds to the weak learner a reweighed version of $S$. Denote the $m-1$-dimensional simplex by $\Delta_m = \{\mathbf{w} \in \mathbb{R}^m \mid \sum_{i\in[m]} w_i = 1, \forall i \in [m], w[i] \geq 0\}$. For a vector $\mathbf{w} \in \Delta_m$, $S_\mathbf{w} = \{(w[i], x_i, y_i)\}_{i=1}^m$ is the sample $S$ reweighed by $\mathbf{w}$. The Boosting algorithm runs in $k$ rounds. On round $t$ it sets a weight vector $\mathbf{w}_t \in \Delta_m$, calls the weak learner with input $S_{\mathbf{w}_t}$, and receives a hypothesis $h_t \in H$ as output from the weak learner. After $k$ rounds, the Boosting algorithm returns a classifier $f_\circ : A \to [-1,+1]$, which is a linear combination of the hypotheses received from the weak learner: $f_\circ = \sum_{t\in[k]} \alpha_t h_t$, where $\alpha_1, \ldots, \alpha_k \in \mathbb{R}$.

The literature offers plenty of Boosting algorithms with desirable properties. For concreteness, we use the algorithm `AdaBoost`$^*$ (Rätsch and Warmuth, 2005), since it provides suitable guarantees on the *margin* of its output classifier. For a labeled example $(x, y)$, the quantity $yf_\circ(x)$ is the margin of $f_\circ$ when classifying $x$. If the margin is positive, then $\mathrm{sign} \circ f_\circ$ classifies $x$ correctly. The margin of any function $f$ on a labeled sample $S = \{(x_i, y_i)\}_{i=1}^m$ is defined as

$$M(f, S) = \min_{i\in[m]} y_i f(x_i).$$

If $M(f, S)$ is positive, then the entire sample is classified correctly by $\mathrm{sign} \circ f$.

If $S$ is an i.i.d. sample drawn from a distribution on $A \times \{-1,+1\}$, then classification error of $f_\circ$ on the distribution can be bounded based on $M(f_\circ, S)$ and the pseudo-dimension $d$ of the hypothesis class $H$. The following bound (Schapire and Singer, 1999, Theorem 8) holds with probability $1-\delta$ over the training samples, for any $m \geq d$:

$$\mathbb{P}[Y \cdot f_\circ(X) \leq 0] \leq O\left(\sqrt{\frac{d\ln^2(m/d)/M^2(f_\circ, S) + \ln(1/\delta)}{m}}\right). \tag{24}$$

In fact, inspection of the proof of this bound in Schapire and Singer (1999) reveals that the only property of the hypothesis class $H$ that is used to achieve this result is the following bound, due to Haussler and Long (1995), on the covering number of a hypothesis class $H$ with pseudo-dimension $d$:

$$\forall \gamma \in (0,1], \quad \mathcal{N}_m(\gamma, \mathcal{H}, \infty) \leq \left(\frac{em}{\gamma d}\right)^d. \tag{25}$$

Thus, Eq. (24) holds whenever this covering bound holds—a fact that will be useful to us.

For `AdaBoost`$^*$, a guarantee on the size of the margin of $f_\circ$ can be achieved if one can provide a guarantee on the *edge* of the hypotheses returned by the weak learner. The edge of a hypothesis measures of how successful it is in classifying labeled examples. Let $h : A \to [-1,+1]$ be a hypothesis and let $D$ be a distribution over $A \times \{-1,+1\}$. The edge of $h$ with respect to $D$ is

$$\Gamma(h, D) \triangleq \mathbb{E}_{(X,Y)\sim D}[Y \cdot h(X)].$$

For a weighted and labeled sample $S = \{(w_i, x_i, y_i)\}_{i\in[m]} \subseteq \mathbb{R}_+ \times A \times \{-1,+1\}$,

$$\Gamma(h, S) \triangleq \sum_{i\in[m]} w_i y_i h(x_i).$$

Note that if $h(x)$ is interpreted as the probability of $h$ to emit 1 for input $x$, then $\frac{1-\Gamma(h,D)}{2}$ is the expected misclassification error of $h$ on $D$. Thus, a positive edge implies a labeling success of more than chance. For AdaBoost$^*$, a positive edge on each of the weighted samples fed to the weak learner suffices to guarantee a positive margin of its output classifier $f_\circ$.

**Theorem 21 (Rätsch and Warmuth 2005)** *Assume* AdaBoost$^*$*receives a labeled sample $S$ of size $m$ as input. Suppose that* AdaBoost$^*$ *runs for $k$ rounds and returns the classifier $f_\circ$. If for every round $t \in [k]$, $\Gamma(h_t, S_{\mathbf{w}_t}) \geq \rho$, then $M(f_\circ, S) \geq \rho - \sqrt{2 \ln m/k}$.*

We present a simple corollary, which we will use when analyzing Boosting for MIL. This corollary shows that AdaBoost$^*$ can be used to transform a weak learner that approximates the best edge of a weighted sample to a Boosting algorithm that approximates the best margin of a labeled sample. The proof of the corollary employs the following well known result, originally by von Neumann (1928) and later extended (see e.g. Nash and Sofer, 1996). For a hypothesis class $H$, denote by $\mathrm{co}(H)$ the set of all linear combinations of hypotheses in $H$. We say that $H \subseteq [-1,+1]^A$ is compact with respect to a sample $S = \{(x_i, y_i)\}_{i\in[m]} \subseteq A \times \{-1,+1\}$ if the set of vectors $\{(h(x_1), \ldots, h(x_m)) \mid h \in H\}$ is compact.

**Theorem 22 (The Strong Min-Max theorem)** *If $H$ is compact with respect to $S$, then*

$$\min_{\mathbf{w}\in\Delta_m} \sup_{h\in H} \Gamma(h, S_{\mathbf{w}}) = \sup_{f\in \mathrm{co}(H)} M(f, S).$$

**Corollary 23** *Suppose that* AdaBoost$^*$ *is executed with an input sample $S$, and assume that $H$ is compact with respect to $S$. Assume the weak learner used by* AdaBoost$^*$ *has the following guarantee: For any $\mathbf{w} \in \Delta_m$, if the weak learner receives $S_{\mathbf{w}}$ as input, then with probability at least $1-\delta$ it returns a hypothesis $h_\circ$ such that*

$$\Gamma(h_\circ, S_{\mathbf{w}}) \geq g(\sup_{h\in H} \Gamma(h, S_{\mathbf{w}})),$$

*where $g : [-1,+1] \to [-1,+1]$ is some fixed non-decreasing function. Then for any input sample $S$, if* AdaBoost$^*$ *runs $k$ rounds, it returns a linear combination of hypotheses $f_\circ = \sum_{t\in[k]} \alpha_t h_t$, such that with probability at least $1-k\delta$*

$$M(f_\circ, S) \geq g(\sup_{f\in\mathrm{co}(H)} M(f, S)) - \sqrt{2\ln m/k}.$$

**Proof** By Theorem 22, $\min_{\mathbf{w}\in\Delta_m} \sup_{h\in H} \Gamma(h, S_{\mathbf{w}}) = \sup_{f\in\mathrm{co}(H)} M(f, S)$. Thus, for any vector of weights $\mathbf{w}$ in the simplex, $\sup_{h\in H} \Gamma(h, S_{\mathbf{w}}) \geq \sup_{f\in\mathrm{co}(H)} M(f, S)$. It follows that in each round, the weak learner that receives $S_{\mathbf{w}_t}$ as input returns a hypothesis $h_t$ such that $\Gamma(h_t, S_{\mathbf{w}_t}) \geq g(\sup_{h\in H} \Gamma(h, S_{\mathbf{w}_t})) \geq g(\sup_{f\in\mathrm{co}(H)} M(f, S))$. By Theorem 21, it follows that $M(f_\circ, S) \geq g(\sup_{f\in\mathrm{co}(H)} M(f, S)) - \sqrt{2\ln m/k}$. ■

### 7.2 The Weak Learner

In this section we will present our weak learner for MIL and provide guarantees for the edge it achieves. Our guarantees depend on boundedness properties of the bag-function $\psi$, which we define below. To motivate our definition of boundedness, consider the $p$-norm bag functions (see Def. 3), defined by $\psi_p(\mathbf{z}) \triangleq \left(\frac{1}{n}\sum_{i=1}^{n}(z[i]+1)^p\right)^{1/p} - 1$. Recall that this class of functions includes the $\max$ function ($\psi_\infty$) and the average function ($\psi_1$) as two extremes. Assume $R \subseteq [r]$ for some $r \in \mathbb{N}$. It is easy to verify that for any natural $n$, any sequence $z_1, \ldots, z_n \in [-1,+1]$, and all $p \in [1,\infty]$,

$$\frac{1}{n}\sum_{i\in[n]} z_i \leq \psi_p(z_1,\ldots,z_n) \leq \sum_{i\in[n]} z_i + n - 1.$$

Since $R \subseteq [r]$, it follows that for all $(z_1,\ldots,z_n) \in [-1,+1]^{(R)}$,

$$\frac{1}{r}\sum_{i\in[n]} z_i \leq \psi_p(z_1,\ldots,z_n) \leq \sum_{i\in[n]} z_i + r - 1. \tag{26}$$

We will show that in cases where the bag function is linearly bounded in the sum of its arguments, as in Eq. (26), a single-instance learning algorithm can be used to learn MIL. Our weak learner will be parameterized by the boundedness parameters of the bag-function, defined formally as follows.

**Definition 24** *A function* $\psi : [-1,+1]^{(R)} \to [-1,+1]$ *is* $(a,b,c,d)$-bounded *if for all* $(z_1,\ldots,z_n) \in [-1,+1]^{(R)}$,

$$a\sum_{i\in[n]} z_i + b \leq \psi(z_1,\ldots,z_n) \leq c\sum_{i\in[n]} z_i + d.$$

Thus, for all $p \in [1,\infty)$, $\psi_p$ over bags of size at most $r$ is $(\frac{1}{r},0,1,r-1)$-bounded.

Before listing the weak learner `MILearn`, we introduce some notations. $\mathbf{h}_{\mathrm{pos}}$ denotes a special bag-hypothesis that labels all bags as $+1$: $\forall x \in \mathcal{X}^{(R)}, \quad \mathbf{h}_{\mathrm{pos}}(x) = 1$. We denote $\overline{\mathcal{H}}_+ \triangleq \overline{\mathcal{H}} \cup \{\mathbf{h}_{\mathrm{pos}}\}$. Let $\mathcal{A}$ be an algorithm that receives a labeled and weighted instance sample as input, and returns a hypothesis $h \in \mathcal{H}$. The result of running $\mathcal{A}$ with input $S$ is denoted $\mathcal{A}(S) \in \mathcal{H}$.

The algorithm `MILearn`, listed as Algorithm 1 below, accepts as input a bag sample $\overline{S}$ and a bounded bag-function $\psi$. It also has access to the algorithm $\mathcal{A}$. We sometimes emphasize that `MILearn` uses a specific algorithm $\mathcal{A}$ as an oracle by writing $\texttt{MILearn}^{\mathcal{A}}$. `MILearn` constructs a sample of instances $S_I$ from the instances that make up the bags in $\overline{S}$, labeling each instance in $S_I$ with the label of the bag it came from. The weights of the instances depend on whether the bag they came from was positive or negative, and on the boundedness properties of $\psi$. Having constructed $S_I$, `MILearn` calls $\mathcal{A}$ with $S_I$. It then decides whether to return the bag-hypothesis induced by applying $\psi$ to $\mathcal{A}(S_I)$, or to simply return $\mathbf{h}_{\mathrm{pos}}$.

It is easy to see that the time complexity of `MILearn` is bounded by $O(f(N)+N)$, where $N$ is the total number of instances in the bags of $\overline{S}$, and $f(n)$ is an upper bound on the time complexity of $\mathcal{A}$ when running on a sample of size $n$. As we presently show, the output of `MILearn` is a bag-hypothesis in $\overline{\mathcal{H}}_+$ whose edge on $\overline{S}$ depends on the best achievable edge for $\overline{S}$.

The guarantees for $\texttt{MILearn}^{\mathcal{A}}$ depend on the properties of $\mathcal{A}$. We define two properties that we consider for $\mathcal{A}$. The first property is that the edge of the hypothesis $\mathcal{A}$ returns is close to the best possible one on the input sample.

---

**Algorithm 1:** `MILearn`$^{\mathcal{A}}$

---

**Assumptions**:

- $\mathcal{H} \in [-1, +1]^{\mathcal{X}}$
- Algorithm $\mathcal{A}$ receives a weighted instance sample and returns a hypothesis in $\mathcal{H}$.

**Input**:

- $\overline{S} \triangleq \{(w_i, \bar{\mathbf{x}}_i, y_i)\}_{i\in[m]}$ — a labeled and weighted sample of bags,
- $\psi$ — an $(a, b, c, d)$-bounded bag-function.

**Output**: $h_\circ \in \overline{\mathcal{H}}_+$.

$\alpha_{(+1)} \leftarrow a$, $\alpha_{(-1)} \leftarrow c$.

$S_I \leftarrow \{(\alpha_{y_i} \cdot w_i, x_i[j], y_i)\}_{i\in[m], j\in[r]}$.

$h_I \leftarrow \mathcal{A}(S_I)$.

**if** $\Gamma(\overline{h}_I, \overline{S}) \geq \Gamma(\mathbf{h}_{\mathrm{pos}}, \overline{S})$ **then**

  $h_\circ \leftarrow \overline{h}_I$,

**else**

  $h_\circ \leftarrow \mathbf{h}_{\mathrm{pos}}$.

Return $h_\circ$.

---

**Definition 25 ($\epsilon$-optimal)** *An algorithm $\mathcal{A}$ that accepts a weighted and labeled sample of instances in $\mathcal{X}$ and returns a hypothesis in $\mathcal{H}$ is* $\epsilon$-optimal *if for all weighted samples $S \subseteq \mathbb{R}_+ \times \mathcal{X} \times \{-1, +1\}$ with total weight $W$,*

$$\Gamma(\mathcal{A}(S), S) \geq \sup_{h\in\mathcal{H}} \Gamma(h, S) - \epsilon W.$$

The second property is that the edge of the hypothesis that $\mathcal{A}$ returns is close to the best possible one on the input sample, but only compared to the edges that can be achieved by hypotheses that label all the negative instances of $S$ with $-1$. For a hypothesis class $\mathcal{H}$ and a distribution $D$ over labeled examples, we denote the set of hypotheses in $\mathcal{H}$ that label all negative examples in $D$ with $-1$, by

$$\Omega(\mathcal{H}, D) = \{h \in \mathcal{H} \mid \mathbb{P}_{(X,Y)\sim D}[h(X) = -1 \mid Y = -1] = 1\}.$$

For a labeled sample $S$, $\Omega(\mathcal{H}, S) \triangleq \Omega(\mathcal{H}, U_S)$ where $U_S$ is the uniform distribution over the examples in $S$.

**Definition 26 (one-sided-$\epsilon$-optimal)** *An algorithm $\mathcal{A}$ that accepts a weighted and labeled sample of instances in $\mathcal{X}$ and returns a hypothesis in $\mathcal{H}$ is* one-sided-$\epsilon$-optimal *if for all weighted samples $S \subseteq \mathbb{R}_+ \times \mathcal{X} \times \{-1, +1\}$ with total weight $W$,*

$$\Gamma(\mathcal{A}(S), S) \geq \sup_{h\in\Omega(\mathcal{H},S)} \Gamma(h, S) - \epsilon W.$$

Clearly, any algorithm which is $\epsilon$-optimal is also one-sided-$\epsilon$-optimal, thus the first requirement from $\mathcal{A}$ is stronger. In our results below we compare the edge achieved using `MILearn` to the best possible edge for the sample $\overline{S}$. Denote the best edge achievable for $\overline{S}$ by a hypothesis in $\mathcal{H}$ by

$$\gamma^* \triangleq \sup_{h \in \overline{\mathcal{H}}} \Gamma(h, \overline{S}).$$

We denote by $\gamma^*_+$ the best edge that can be achieved by a hypothesis in $\Omega(\overline{\mathcal{H}}, \overline{S})$. Formally,

$$\gamma^*_+ \triangleq \sup_{h \in \Omega(\overline{\mathcal{H}}, \overline{S})} \Gamma(h, \overline{S}).$$

Denote the weight of the positive bags in the input sample $\overline{S}$ by $W_+ = \sum_{i:y_i=+1} w_i$ and the weight of the negative bags by $W_- = \sum_{i:y_i=-1} w_i$. We will henceforth assume without loss of generality that the total weight of all bags in the input sample is $1$, that is $W_+ + W_- = 1$.

Note that for any $(a,b,c,d)$-bounded $\psi$, if there exists any sequence $z_1, \ldots, z_n$ such that $\psi(z_1, \ldots, z_n) = -1$, then

$$a \sum_{i \in [n]} z_i + b \leq -1 \leq c \sum_{i \in [n]} z_i + d. \tag{27}$$

This implies

$$\frac{-1-d}{c} \leq \sum_{i \in [n]} z_i \leq \frac{-1-b}{a}.$$

Rearranging, we get $d - \frac{c}{a}b - \frac{c}{a} + 1 \geq 0$, with equality if Eq. (27) holds with equalities. The next theorem provides a guarantee for `MILearn` that depends on the tightness of this inequality for the given bag function. As evident from Theorem 21, to guarantee a positive margin for the output of `AdaBoost`$^*$ when used with `MILearn` as the weak learner, we need to guarantee that the edge of the hypothesis returned by `MILearn` is always positive. Since the best edge cannot be more than $1$, we emphasize in the theorem below that the edge achieved by `MILearn` is positive at least when the best edge is $1$ (and possibly also for smaller edges, depending on the parameters). We subsequently show how these general guarantees translate to a specific result for the $\max$ function, and other bag functions with the same boundedness properties.

**Theorem 27** *Let $r \in \mathbb{N}$ and $R \subseteq [r]$. Let $\psi : [-1,+1]^{(R)} \to [-1,+1]$ be an $(a,b,c,d)$-bounded bag-function such that $0 < a \leq c$. Let $\epsilon \in [0, \frac{1}{rc})$, and assume that $d - \frac{c}{a}b - \frac{c}{a} + 1 = \eta$. Denote $Z = \frac{c}{a}$. Consider running the algorithm `MILearn`$^{\mathcal{A}}$ with a weighted bag sample $\overline{S}$ of total weight $1$, and let $h_\circ$ be the hypothesis returned by `MILearn`$^{\mathcal{A}}$. Then*

1. *If $\mathcal{A}$ is $\epsilon$-optimal then*

$$\Gamma(h_\circ, \overline{S}) \geq \frac{Z\gamma^* - Z + \frac{1}{Z} - \frac{\eta}{2}(1 + \frac{1}{Z}) - rc\epsilon}{1 + (1 - \frac{\eta}{2})(1 - \frac{1}{Z})}.$$

*Thus, $\Gamma(h_\circ, \overline{S}) > 0$ whenever*

$$\gamma^* > 1 - \frac{1}{Z^2} + \frac{\eta}{2}(\frac{1}{Z} + \frac{1}{Z^2}) + \frac{rc\epsilon}{Z}.$$

*In particular, if $\eta \leq 2(1 - rc\epsilon)/(Z+1)$ and $\gamma^* = 1$ then $\Gamma(h_\circ, \overline{S}) > 0$.*

2. *If $\mathcal{A}$ is one-sided-$\epsilon$-optimal, and $\psi(z_1, \ldots, z_n) = -1$ only if $z_1 = \ldots = z_n = -1$, then*

$$\Gamma(h_\circ, \overline{S}) \geq \frac{\gamma_+^* - \frac{\eta}{2}(Z+1) - rc\epsilon Z}{2Z - 1 - \frac{\eta}{2}(Z-1)}.$$

*Thus, $\Gamma(h_\circ, \overline{S}) > 0$ whenever*

$$\gamma_+^* > \frac{\eta}{2}(Z+1) + rc\epsilon Z.$$

*In particular, if $\eta \leq 2(1 - rc\epsilon Z)/(Z+1)$ and $\gamma_+^* = 1$ then $\Gamma(h_\circ, \overline{S}) > 0$.*

The proof of the theorem is provided in Appendix A. This theorem is stated in general terms, as it holds for any bounded $\psi$. In particular, if $\psi$ is any function between an average and a $\max$, including any of the $p$-norm bag functions $\psi_p$ defined in Def. 3, we can simplify the result, as captured by the following corollary.

**Corollary 28** *Let $\mathcal{H} \subseteq [-1,+1]^{\mathcal{X}}$. Let $R \subseteq [r]$, and $\epsilon \in [0, \frac{1}{r})$. Assume a bag function $\psi : [-1,+1]^{(R)} \to [-1,+1]$ such that for any $z_1, \ldots, z_n \in [-1,+1]$,*

$$\frac{1}{n}\sum_{i \in [n]} z_i \leq \psi(z_1, \ldots, z_n) \leq \max_{i \in [n]} z_i.$$

*Let $h_\circ$ be the hypothesis returned by* $\texttt{MILearn}^{\mathcal{A}}$*. Then*

1. *If $\mathcal{A}$ is $\epsilon$-optimal for some $\epsilon \in [0, 1/r]$, then*

$$\Gamma(h_\circ, \overline{S}) \geq \frac{r^2\gamma^* + 1 - r^2(1+\epsilon)}{2r - 1}.$$

*Thus $\Gamma(h_\circ, \overline{S}) > 0$ whenever $\gamma^* \geq 1 - \frac{1}{r^2} + \frac{\epsilon}{r}$. In particular, if $\gamma^* = 1$ then $\Gamma(h_\circ, \overline{S}) > 0$.*

2. *If $\mathcal{A}$ is one-sided-$\epsilon$-optimal some $\epsilon \in [0, 1/r^2]$, then*

$$\Gamma(h_\circ, \overline{S}) \geq \frac{\gamma_+^* - r^2\epsilon}{2r - 1}.$$

*Thus $\Gamma(h_\circ, \overline{S}) > 0$ whenever $\gamma_+^* > r^2\epsilon$. In particular, if $\gamma_+^* = 1$ then $\Gamma(h_\circ, \overline{S}) > 0$.*

**Proof** Let $z_1, \ldots, z_n \in [-1,+1]$. We have

$$\max_{i \in [n]} z_i \leq \sum_{i \in [n]} z_i - (n-1)\min(z_i) \leq \sum_{i \in [n]} z_i + n - 1.$$

Therefore, by the assumption on $\psi$, for any $n \in R$

$$\psi(z_1, \ldots, z_n) \leq \sum_{i \in [n]} z_i + n - 1 \leq \sum_{i \in [n]} z_i + r - 1.$$

In addition

$$\frac{1}{r}\sum_{i \in [n]} z_i \leq \frac{1}{n}\sum_{i \in [n]} z_i \leq \psi(z_1, \ldots, z_n).$$

Therefore $\psi$ is $(\frac{1}{r}, 0, 1, r-1)$-bounded. It follows that $Z = r$ in this case, and $d - Zb - Z + 1 = 0$. Claim (1) follows by applying case (1) of Theorem 27 with $\eta = 0$.

For claim (2) we apply case (2) of Theorem 27. Thus we need to show that if $\psi(z_1, \ldots, z_n) = -1$ and $z_1, \ldots, z_n \in [-1, +1]$, then $z_1 = \ldots = z_n = -1$. We have that

$$-1 \leq \frac{1}{n} \sum_{i \in [n]} z_i \leq \psi(z_1, \ldots, z_n) \leq -1.$$

Therefore $\frac{1}{n} \sum_{i \in [n]} z_i = -1$. Since no $z_i$ can be smaller than $-1$, $z_1 = \ldots = z_n = -1$. Thus case (2) of Theorem 27 holds. We get our claim (2) directly by subsituting the boundedness parameters of $\psi$ in Theorem 27 case (2). ■

### 7.3 From Single-Instance Learning to Multi-Instance Learning

In this section we combine the guarantees on `MILearn` with the guarantees on `AdaBoost`$^*$, to show that efficient agnostic PAC-learning of the underlying instance hypothesis $\mathcal{H}$ implies efficient PAC-learning of MIL. For simplicity we formalize the results for the natural case where the bag function is $\psi = \max$. Results for other bounded bag functions can be derived in a similar fashion.

First, we formally define the notions of agnostic and one-sided PAC-learning algorithms. We then show that given an algorithm on instances that satisfies one of these definitions, we can construct an algorithm for MIL which approximately maximizes the margin on an input bag sample. Specifically, if the input bag sample is realizable by $\overline{\mathcal{H}}$, then the MIL algorithm we propose will find a linear combination of bag hypotheses that classifies the sample with zero error, and with a positive margin. Combining this with the margin-based generalization guarantees mentioned in Section 7.1, we conclude that we have an efficient PAC-learner for MIL.

**Definition 29 (Agnostic PAC-learner and one-sided PAC-learner)** *Let $\mathcal{B}(\epsilon, \delta, S)$ be an algorithm that accepts as input $\delta, \epsilon \in (0,1)$, and a labeled sample $S \in (\mathcal{X} \times \{-1,+1\})^m$, and emits as output a hypothesis $h \in \mathcal{H}$. $\mathcal{B}$ is an* agnostic PAC-learner *for $\mathcal{H}$ with complexity $c(\epsilon, \delta)$ if $\mathcal{B}$ runs for no more than $c(\epsilon, \delta)$ steps, and for any probability distribution $D$ over $\mathcal{X} \times \{-1,+1\}$, if $S$ is an i.i.d. sample from $D$ of size $c(\epsilon, \delta)$, then with probability at least $1-\delta$ over $S$ and the randomization of $\mathcal{B}$,*

$$\Gamma(\mathcal{B}(\epsilon, \delta, S), D) \geq \sup_{h \in \mathcal{H}} \Gamma(h, D) - \epsilon.$$

*$\mathcal{B}$ is a* one-sided PAC-learner *if under the same conditions, with probability at least $1 - \delta$*

$$\Gamma(\mathcal{B}(\epsilon, \delta, S), D) \geq \sup_{h \in \Omega(\mathcal{H}, D)} \Gamma(h, D) - \epsilon.$$

Given an agnostic PAC-learner $\mathcal{B}$ for $\mathcal{H}$ and parameters $\epsilon, \delta \in (0,1)$, the algorithm $\mathcal{O}^{\mathcal{B}}_{\epsilon,\delta}$, listed above as Algorithm 2, is an $\epsilon$-optimal algorithm with probability $1 - \delta$. Similarly, if $\mathcal{B}$ is a one-sided PAC-learner, then $\mathcal{O}^{\mathcal{B}}_{\epsilon,\delta}$ is a one-sided-$\epsilon$-optimal algorithm with probability $1 - \delta$. Our MIL algorithm is then simply `AdaBoost`$^*$ with `MILearn`$^{\mathcal{O}^{\mathcal{B}}_{\epsilon,\delta}}$ as the (high probability) weak learner. It is easy to see that this algorithm learns a linear combination of hypotheses from $\overline{\mathcal{H}}_+$. We also show below that under certain conditions this linear combination induces a positive margin on the input

**Algorithm 2:** $\mathcal{O}^{\mathcal{B}}_{\epsilon,\delta}$

**Assumptions**:

- $\epsilon, \delta \in (0, 1)$.
- $\mathcal{B}$ receives a labeled instance sample as input and returns a hypothesis in $\mathcal{H}$.
- Algorithm $\mathcal{B}$ is a one-sided (or agnostic) PAC-learning algorithm with complexity $c(\epsilon, \delta)$.

**Input**: A labeled and weighted instance sample
$S = \{(w_i, x_i, y_i)\}_{i\in[m]} \subseteq \mathbb{R}_+ \times \mathcal{X} \times \{-1, +1\}$.

**Output**: A hypothesis in $\mathcal{H}$

For all $i \in [m]$, $p_i \leftarrow w_i / \sum_{i\in[m]} w_i$.
For each $t \in [c(\epsilon, \delta)]$, independently draw a random $j_t$ such that $j_t = i$ with probability $p_i$.
$\tilde{S} \leftarrow \{(x_{j_t}, y_{j_t})\}_{t\in[c(\epsilon,\delta)]}$.
$h \leftarrow \mathcal{B}(\tilde{S})$
Return $h$.

bag sample with high probability. Given this guaranteed margin, we bound the generalization error of the learning algorithm via Eq. (24).

The computational complexity of $\mathcal{O}^{\mathcal{B}}_{\epsilon,\delta}$ is polynomial in $c(\epsilon, \delta)$ and in the instance-sample size $m$. Therefore, the computational complexity of $\texttt{MILearn}^{\mathcal{O}^{\mathcal{B}}_{\epsilon,\delta}}$ is polynomial in $c(\epsilon, \delta)$ and in $N$, where $N$ is the total number of instances in the input bag sample $\overline{S}$.

For 1-Lipschitz bag functions which have desired boundedness properties, both the sample complexity and the computational complexity of the proposed MIL algorithm are polynomial in the maximal bag size and linear in the complexity of the underlying instance hypothesis class. This is formally stated in the following theorem, for the case of a realizable distribution over labeled bags. Note that in particular, the theorem holds for all the $p$-norm bag-functions, since they are 1-Lipschitz and satisfy the boundedness conditions.

**Theorem 30** *Let $\mathcal{H} \subseteq [-1, +1]^{\mathcal{X}}$ be a hypothesis class with pseudo-dimension $d$. Let $\mathcal{B}$ be a one-sided PAC-learner for $\mathcal{H}$ with complexity $c(\epsilon, \delta)$. Let $r \in \mathbb{N}$, and let $R \subseteq [r]$. Assume that the bag function $\psi : [-1, +1]^{(R)} \to [-1, +1]$ is 1-Lipschitz with respect to the infinity norm, and that for any $(z_1, \ldots, z_n) \in [-1, +1]^{(R)}$*

$$\frac{1}{n} \sum_{i\in[n]} z_i \leq \psi(z_1, \ldots, z_n) \leq \max_{i\in[n]} z_i.$$

*Assume that $\overline{\mathcal{H}}$ is compact with respect to any sample of size $m$. Let $D$ be a distribution over $\mathcal{X}^{(R)} \times \{-1, +1\}$ which is realizable by $\overline{\mathcal{H}}$, that is there exists an $h \in \mathcal{H}$ such that $\mathbb{P}_{(\bar{\mathbf{X}},Y)\sim D}[\overline{h}(\bar{\mathbf{X}}) = Y] = 1$. Assume $m \geq 10d \ln(er)$, and let $\epsilon = \frac{1}{2r^2}$ and $k = 32(2r - 1)^2 \ln(m)$.*

*For all $\delta \in (0, 1)$, if $\texttt{AdaBoost}^*$ is executed for $k$ rounds on a random sample $S \sim D^m$, with $\texttt{MILearn}^{\mathcal{O}^{\mathcal{B}}_{\epsilon,\delta/2k}}$ as the weak learner, then with probability $1 - \delta$, the classifier $f_\circ$ returned by*

`AdaBoost`$^*$ *satisfies*

$$\mathbb{P}_D[Yf(\bar{\mathbf{X}}) \leq 0] \leq O\left(\sqrt{\frac{dr^2\ln(r)\ln^2(m) + \ln(2/\delta)}{m}}\right). \quad (28)$$

**Proof** Since $\mathcal{B}$ is a one-sided PAC-learning algorithm, $\mathcal{O}^{\mathcal{B}}_{\epsilon,\delta/2k}$ is one-sided-$\epsilon$-optimal with probability at least $1-\delta/2k$. Therefore, by case (2) of Cor. 28, if $\mathtt{MILearn}^{\mathcal{O}^{\mathcal{B}}_{\epsilon,\delta/k}}$ receives a weighted bag sample $S_{\mathbf{w}}$, with probability $1-\delta/2k$ it returns a bag hypothesis $h_\circ \in \overline{\mathcal{H}}_+$ such that

$$\Gamma(h_\circ, S_{\mathbf{w}}) \geq \frac{\sup_{h\in\Omega(\overline{\mathcal{H}},S)}\Gamma(h,S_{\mathbf{w}}) - r^2\epsilon}{2r-1}.$$

Thus, by Cor. 23, if `AdaBoost`$^*$ runs for $k$ rounds then with probability $1-\delta/2$ it returns a linear combination of hypotheses from $\overline{\mathcal{H}}_+$ such that

$$M(f_\circ, S) \geq \frac{\sup_{f\in\mathrm{co}(\Omega(\overline{\mathcal{H}},S))} M(f,S) - r^2\epsilon}{2r-1} - \sqrt{2\ln m/k}. \quad (29)$$

Due to the realizability assumption for $D$, there is an $h \in \Omega(\overline{\mathcal{H}}, S)$ that classifies correctly the bag sample $S$. It follows that for any weighting $\mathbf{w} \in \Delta_m$ of $S$, $\Gamma(h, S_{\mathbf{w}}) = 1$. It is easy to verify that since $\overline{\mathcal{H}}$ is compact with respect to $S$, then so is $\Omega(\mathcal{H}, S)$. Thus, by Theorem 22, $\sup_{f\in\mathrm{co}(\Omega(\overline{\mathcal{H}},S))} M(f,S) = \min_{\mathbf{w}}\sup_{h\in\Omega(\overline{\mathcal{H}},S)}\Gamma(h,S_{\mathbf{w}}) = 1$. Substituting $\epsilon$ and $k$ with their values, setting $\sup_{f\in\mathrm{co}(\Omega(\overline{\mathcal{H}},S))} M(f,S) = 1$ in Eq. (29) and simplifying, we get that with probability $1-\delta/2$

$$M(f_\circ, S) \geq \frac{1}{8r-4}. \quad (30)$$

We would now like to apply the generalization bound in Eq. (24), but for this we need to show that Eq. (25) holds for $\overline{\mathcal{H}}$. We have the following bound on the covering numbers of $\overline{\mathcal{H}}$, for all $\gamma \in (0,1]$:

$$\mathcal{N}_m(\gamma, \overline{\mathcal{H}}, \infty) \leq \mathcal{N}_{rm}(\gamma, \mathcal{H}, \infty) \leq \left(\frac{erm}{\gamma d}\right)^d.$$

The first inequality is due to Cor. 13 and the fact that $\psi$ is 1-Lipschitz, and the second inequality is due to Haussler and Long (1995) and the pseudo-dimension of $\mathcal{H}$ (see Eq. (25) above). This implies

$$\mathcal{N}_m(\gamma, \overline{\mathcal{H}}, \infty) \leq \left(\frac{erm}{\gamma d}\right)^d = \left(\frac{em}{\gamma d}\right)^d \cdot e^{d\ln(r)} = \left(\frac{em}{\gamma\cdot 10d\ln(er)}\right)^d \cdot (10\ln(er))^d e^{d\ln(r)}$$
$$= \left(\frac{em}{\gamma\cdot 10d\ln(er)}\right)^d \cdot e^{d(\ln(10\ln(er))+\ln(r))}.$$

Therefore, for $m \geq 10d\ln(er)$

$$\mathcal{N}_m(\gamma, \overline{\mathcal{H}}_+, \infty) \leq 1 + \mathcal{N}_m(\gamma, \overline{\mathcal{H}}, \infty) \leq 1 + \left(\frac{em}{\gamma\cdot 10d\ln(er)}\right)^d \cdot e^{d(\ln(10\ln(er))+\ln(r))}$$
$$\leq \left(\frac{em}{\gamma\cdot 10d\ln(er)}\right)^d \cdot e^{d(\ln(10\ln(er))+\ln(er))}.$$

Now, $\ln(10\ln(er)) + \ln(er) = \ln(10) + \ln(\ln(er)) + \ln(er) \leq \ln(10) + 2\ln(er) \leq 3 + 2\ln(er) \leq 5\ln(er)$. Therefore,

$$\mathcal{N}_m(\gamma, \overline{\mathcal{H}}_+, \infty) \leq \left(\frac{em}{\gamma \cdot 10d\ln(er)}\right)^d \cdot e^{5d\ln(er)} \leq \left(\frac{e^2 m}{\gamma \cdot 10d\ln(er)}\right)^{5d\ln(er)} \leq \left(\frac{em}{\gamma \cdot 10d\ln(er)}\right)^{10d\ln(er)}.$$

Thus, for $m \geq 10d\ln(er)$, Eq. (25) holds for $\overline{\mathcal{H}}_+$ when substituting $d$ with $d_r = 10d\ln(er)$. This means the generalization bound in Eq. (24) holds when substituting $d$ with $d_r$ as well. It follows that with probability $1 - \delta/2$

$$\mathbb{P}[Yf_\circ(X) \leq 0] \leq O\left(\sqrt{\frac{d_r \ln^2(m/d_r)/M^2(f_\circ, S) + \ln(1/\delta)}{m}}\right).$$

Now, with probability $1 - \delta/2$, by Eq. (30) we have $M(f_\circ, S) \geq 1/(8r-4)$. Combining the two inequalities and applying the union bound, we have that with probability $1 - \delta$

$$\begin{aligned}\mathbb{P}[Yf_\circ(X) \leq 0] &\leq O\left(\sqrt{\frac{d_r(8r-4)^2\ln^2(m/d_r) + \ln(2/\delta)}{m}}\right)\\ &\leq O\left(\sqrt{\frac{10d\ln(er)(8r-4)^2\ln^2(m) + \ln(2/\delta)}{m}}\right).\end{aligned}$$

Due to the O-notation we can simplify the right-hand side to get Eq. (28).

■

Similar generalization results for Boosting can be derived for margin-learning as well, using covering-numbers arguments as discussed in Schapire et al. (1998). The theorem above leads to the following conclusion.

**Corollary 31** *If there exists a one-sided PAC-learning algorithm for $\mathcal{H}$ with polynomial run-time in $\frac{1}{\epsilon}$ and $\frac{1}{\delta}$, then there exists a PAC-learning algorithm for classical MIL on $\mathcal{H}$, which has polynomial run-time in $r$, $\frac{1}{\epsilon}$ and $\frac{1}{\delta}$.*

Cor. 31 is similar in structure to Theorem 1: Both state that if the single-instance problem is solvable with one-sided error, then the realizable MIL problem is solvable. Theorem 1 applies only to bags with statistically independent instances, while Cor. 31 applies to bags drawn from an arbitrary distribution. The assumption of Theorem 1 is similarly weaker, as it only requires that the single-instance PAC-learning algorithm handle random one-sided noise, while Cor. 31 requires that the single-instance algorithm handle arbitrary one-sided noise. Of course, Cor. 31 does not contradict the hardness result provided for APRs in Auer et al. (1998). Indeed, this hardness result states that if there exists a MIL algorithm for $d$-dimensional APRs which is polynomial in both $r$ and $d$, then $\mathcal{RP} = \mathcal{NP}$. Our result does not imply that such an algorithm exists, since there is no known agnostic or one-sided PAC-learning algorithm for APRs which is polynomial in $d$.

**Example: Half-spaces** We have shown a simple and general way, independent of hypothesis class, to create a PAC-learning algorithm for classical MIL from a learning algorithm that runs on single instances. Whenever an appropriate polynomial algorithm exists for the non-MIL learning problem, the resulting MIL algorithm will also be polynomial in $r$. To illustrate, consider for instance the algorithm proposed in Shalev-Shwartz et al. (2010). This algorithm is an agnostic PAC-learner of fuzzy kernelized half-spaces with an $L$-Lipschitz transfer function, for some constant $L > 0$. Its time complexity and sample-complexity are at most $\mathrm{poly}((\frac{L}{\epsilon})^L \cdot \ln(\frac{1}{\delta}))$. Since this complexity bound is polynomial in $1/\epsilon$ and in $1/\delta$, this algorithm can serve as the algorithm $\mathcal{B}$ in Theorem 30, and Cor. 31 holds. Thus, we can generate an algorithm for PAC-learning MIL with complexity that depends directly on the complexity of this learner, and is polynomial in $r$, $\frac{1}{\epsilon}$ and $\frac{1}{\delta}$. The full MIL algorithm for fuzzy kernelized half-spaces can thus be described as follows: Run `AdaBoost*` with the weak learner $\texttt{MILearn}^{\mathcal{O}^{\mathcal{B}}_{\epsilon,\delta}}$, where `MILearn` is listed in Algorithm 1, $\mathcal{O}^{\mathcal{B}}_{\epsilon,\delta}$ is listed in Algorithm 2, and $\mathcal{B}$ is the agnostic PAC-learner from Shalev-Shwartz et al. (2010). The input to `AdaBoost*` is a labeled sample of bags, and the output is a real-valued classifier for bags.

More generally, using the construction we proposed here, any advancement in the development of algorithms for agnostic or one-sided learning of any hypothesis class translates immediately to an algorithm for PAC-learning MIL with the same hypothesis class, and with corresponding complexity guarantees.

## 8. Conclusions

In this work we have provided a new theoretical analysis for Multiple Instance Learning with any underlying hypothesis class. We have shown that the dependence of the sample complexity of generalized MIL on the number of instances in a bag is only poly-logarithmic, thus implying that the statistical performance of MIL is only mildly sensitive to the size of the bag. The analysis includes binary hypotheses, real-valued hypotheses, and margin learning, all of which are used in practice in MIL applications. Our sample complexity results can be summarized as follows, where $d$ is the VC dimension or pseudo-dimension of the underlying hypothesis class, and $r$ is the maximal/average bag size.

- The VC dimension of binary MIL is $O(d\log(r))$.
- For non-trivial bag functions, there are hypothesis classes such that the VC dimension of binary MIL is $\Omega(d\log(r))$.
- The VC dimension of binary MIL with separating hyperplanes in dimension $d$ is $\Omega(d\log(r))$.
- The pseudo-dimension of binary MIL for bag functions that are extensions of monotone Boolean functions is $O(d\log(r))$.
- Covering numbers for MIL hypotheses with Lipschitz bag functions can be bounded by covering numbers for the single instance hypothesis class.
- The fat-shattering dimension of real-valued MIL with Lipschitz bag-functions is poly-logarithmic in the bag size and quasilinear in the fat shattering dimension of the single instance hypothesis class.

- The Rademacher complexity of binary MIL with a bounded average bag size is $O(\sqrt{d\log(r)/m})$ where $m$ is the sample size.
- The Rademacher complexity of real-valued MIL with a Lipschitz loss function and a Lipschitz bag function is upper bounded by a logarithmic dependence on the average bag size and a quasilinear dependence on the Rademacher complexity of the instance hypothesis class.

For classical MIL, where the bag-labeling function is the Boolean OR, and for its natural extension to $\max$, we have presented a new learning algorithm, that classifies bags by executing a learning algorithm designed for single instances. This algorithm provably PAC-learns MIL. In both the sample complexity analysis and the computational analysis, we have shown tight connections between classical supervised learning and Multiple Instance Learning, which holds regardless of the underlying hypothesis class.

Many interesting open problems remain for the generic analysis of MIL. In particular, our results hold under certain assumptions on the bag functions. An interesting open question is whether these assumptions are necessary, or whether useful results can be achieved for other classes of bag functions. Another interesting question is how additional structure within a bag, such as sparsity, may affect the statistical and computational feasibility of MIL. These interesting problems are left for future research.

## Acknowledgments

The authors thank Shai Shalev-Shwartz and Nati Srebro for helpful discussions. Enlightening comments from Phil Long and Sally Goldman on the preliminary version of this work have helped improve the presentation immensely. Sivan Sabato is supported by the Adams Fellowship Program of the Israel Academy of Sciences and Humanities. This work is partly supported by the Gatsby Charitable Foundation.

## References

N. Alon, S. Ben-David, N. Cesa-Bianchi, and D. Haussler. Scale-sensitive dimensions, uniform convergence, and learnability. In *Foundations of Computer Science, 1993. Proceedings., 34th Annual Symposium on*, pages 292–301. IEEE, 1993.

N. Alon, S. Ben-David, N. Cesa-Bianchi, and D. Haussler. Scale-sensitive dimensions, uniform convergence, and learnability. *J. ACM*, 44(4):615–631, 1997.

S. Andrews. *Learning from ambiguous examples*. PhD thesis, Brown University, May 2007.

S. Andrews and T. Hofmann. Multiple-instance learning via disjunctive programming boosting. In *Advances in Neural Information Processing Systems 16*, 2003.

S. Andrews, I. Tsochantaridis, and T. Hofmann. Support vector machines for multiple-instance learning. In *Advances in Neural Information Processing Systems 15*, pages 561–568, 2002.

M. Anthony and P. L. Bartlett. *Neural Network Learning: Theoretical Foundations*. Cambridge University Press, 1999.

P. Auer, P. M. Long, and A. Srinivasan. Approximating hyper-rectangles: learning and pseudorandom sets. *Journal of Computer and System Sciences*, 57(3):376–388, 1998.

B. Babenko, N. Verma, P. Dollar, and S. Belongie. Multiple instance learning with manifold bags. In *Proceedings of the 28th International Conference on Machine Learning (ICML-11)*, ICML '11, pages 81–88, 2011.

P. L. Bartlett and S. Mendelson. Rademacher and Gaussian complexities: Risk bounds and structural results. *Journal of Machine Learning Research*, 3:463–482, 2002.

P. L. Bartlett, S. R. Kulkarni, and S. E. Posner. Covering numbers for real-valued function classes. *IEEE Transactions on Information Theory*, 43(5):1721–1724, 1997.

A. Blum and A. Kalai. A note on learning from multiple-instance examples. *Machine Learning*, 30 (1):23–29, 1998.

C. Cortes and V. Vapnik. Support-vector networks. *Machine Learning*, 20(3):273–297, September 1995.

T. G. Dietterich, R. H. Lathrop, and T. Lozano-Pérez. Solving the multiple instance problem with axis-parallel rectangles. *Artificial Intelligence*, 89(1-2):31–71, 1997.

R. M. Dudley. Central limit theorems for empirical measures. *The Annals of Probability*, 6(6): 899–929, 1978.

R.M. Dudley. The sizes of compact subsets of hilbert space and continuity of gaussian processes. *Journal of Functional Analysis*, 1(3):290 – 330, 1967.

Y. Freund and R.E. Schapire. A decision-theoretic generalization of on-line learning and an application to boosting. *Journal of Computer and System Sciences*, 55(1):119–139, August 1997.

T. Gärtner, P. A. Flach, A. Kowalczyk, and A. J. Smola. Multi-instance kernels. In *Proceedings of the Nineteenth International Conference on Machine Learning*, pages 179–186, 2002.

D. Haussler and P. M. Long. A generalization of Sauer's lemma. *Journal of Combinatorial Theory, Series A*, 71(2):219–240, 1995.

P. M. Long and L. Tan. PAC learning axis-aligned rectangles with respect to product distributions from multiple-instance examples. *Machine Learning*, 30(1):7–21, 1998. ISSN 0885-6125.

O. Maron and T. Lozano-Pérez. A framework for multiple-instance learning. In *Advances in Neural Information Processing Systems 10*, pages 570–576, 1998.

O. Maron and A. L. Ratan. Multiple-instance learning for natural scene classification. In *Proceedings of the Fifteenth International Conference on Machine Learning*, pages 341–349, 1998.

S. Mendelson. Rademacher averages and phase transitions in glivenko-cantelli classes. *IEEE Transactions on Information Theory*, 48(1):251–263, 2002.

S. Nash and A. Sofer. *Linear and Nonlinear Programming*. McGraw-Hill, New York, 1996.

L. Pitt and L. G. Valiant. Computational limitations on learning from examples. Technical report, Harvard University Aiken Computation Laboratory, July 1986.

D. Pollard. *Convergence of Stochastic Processes*. Springer-Verlag, 1984.

L. De Raedt. Attribute-value learning versus inductive logic programming: The missing links (extended abstract). In *Proceedings of the 8th International Workshop on Inductive Logic Programming*, pages 1–8, London, UK, 1998. Springer-Verlag.

G. Rätsch and M. K. Warmuth. Efficient margin maximizing with boosting. *Journal of Machine Learning Research*, 6:2131–2152, December 2005.

S. Sabato and N. Tishby. Homogeneous multi-instance learning with arbitrary dependence. In *Proceedings of the Twenty Second Annual Conference on Computational Learning Theory*, 2009.

S. Sabato, N. Srebro, and N. Tishby. Reducing label complexity by learning from bags. In *Proceedings of the Thirteenth International Conference on Artificial Intelligence and Statistics*, volume 9, pages 685–692, 2010.

N. Sauer. On the density of families of sets. *Journal of Combinatorial Theory Series A*, 13:145–147, 1972.

R. E. Schapire and Y. Singer. Improved boosting algorithms using confidence-rated predictions. *Machine Learning*, 37(3):1–40, 1999.

R.E. Schapire, Y. Freund, P. Bartlett, and W.S. Lee. Boosting the margin: A new explanation for the effectiveness of voting methods. *The Annals of Statistics*, 26(5):1651–1686, October 1998.

S. Shalev-Shwartz, O. Shamir, and K. Learning kernel-based halfspaces with the zero-one loss. In *Proceedings of the Twenty Third Annual Conference on Computational Learning Theory*, 2010.

N. Srebro, K. Sridharan, and A. Tewari. Smoothness, low-noise and fast rates. *CoRR*, abs/1009.3896, 2010.

V. N. Vapnik and A. Ya. Chervonenkis. On the uniform convergence of relative frequencies of events to their probabilities. *Theory of Probability and its applications*, XVI(2):264–280, 1971.

J. von Neumann. Zur theorie der gesellschaftsspiele. *Mathematische Annalen*, 100:295–320, 1928.

N. Weidmann, E. Frank, and B. Pfahringer. A two-level learning method for generalized multi-instance problems. In *Proceedings of the European conference on machine learning*, pages 468–479, 2003.

Q. Zhang and S.A. Goldman. EM-DD: An improved multiple-instance learning technique. In *Advances in Neural Information Processing Systems 14*, 2001.

Z. Zhou, K. Jiang, and M. Li. Multi-instance learning based web mining. *Applied Intelligence*, 22 (2):135–147, 2005.

## Appendix A. Proof of Theorem 27

The first step in providing a guarantee for the edge achieved by `MILearn`, is to prove a guarantee for the edge achieved on the bag sample by the hypothesis returned by $\mathcal{A}$ in step (3) of the algorithm. This is done in the following lemma.

**Lemma 32** *Assume $\psi : [-1,+1]^{(R)} \rightarrow [-1,+1]$ is an $(a,b,c,d)$-bounded bag function with $0 < a \leq c$, and denote $Z = \frac{c}{a}$. Consider running the algorithm `MILearn` with a weighted bag sample $\overline{S}$ of total weight 1. Let $h_I$ be the hypothesis returned by the oracle $\mathcal{A}$ in step (3) of `MILearn`. Let $W$ be the total weight of the sample $S_I$ created in `MILearn`, step (2). Then*

1. *If $\mathcal{A}$ is $\epsilon$-optimal,*

$$\Gamma(\overline{h}_I, \overline{S}) \geq Z\gamma^* + (\frac{1}{Z} - Z + (1 - \frac{1}{Z})(d - Zb))W_+ + Zb - d - \epsilon W.$$

2. *If $\mathcal{A}$ is one-sided-$\epsilon$-optimal, and $\psi(z_1, \ldots, z_n) = -1$ only if $z_1 = \ldots = z_n = -1$, then*

$$\Gamma(\overline{h}_I, \overline{S}) \geq \frac{1}{Z}\gamma_+^* + (\frac{1}{Z} - Z + (1 - \frac{1}{Z})(d - Zb))W_+ + Zb - d + Z - \frac{1}{Z} - \epsilon W.$$

**Proof** For all $h \in \mathcal{H}$, and for all $\bar{\mathbf{x}} = (x_1, \ldots, x_n) \in \mathcal{X}^{(R)}$ we have $\overline{h}(\bar{\mathbf{x}}) = \psi(h(x_1), \ldots, h(x_n))$. Since $\psi$ is $(a,b,c,d)$-bounded, it follows that

$$a \sum_{x \in \bar{\mathbf{x}}} h(x) + b \leq \overline{h}(\bar{\mathbf{x}}) \leq c \sum_{x \in \bar{\mathbf{x}}} h(x) + d. \tag{31}$$

In addition, since $a$ and $c$ are positive we also have

$$(\overline{h}(\bar{\mathbf{x}}) - d)/c \leq \sum_{x \in \bar{\mathbf{x}}} h(x) \leq (\overline{h}(\bar{\mathbf{x}}) - b)/a. \tag{32}$$

Assume the input bag sample is $\overline{S} = \{(w_i, \bar{\mathbf{x}}_i, y_i)\}_{i \in [m]}$. Denote $I_+ = \{i \in [m] \mid y_i = +1\}$ and $I_- = \{i \in [m] \mid y_i = -1\}$. Let $h \in \mathcal{H}$ be a hypothesis. We have

$$\begin{aligned}
\Gamma(\overline{h}, \overline{S}) &= \sum_{i \in I_+} w_i \overline{h}(\bar{\mathbf{x}}_i) - \sum_{i \in I_-} w_i \overline{h}(\bar{\mathbf{x}}_i) \\
&\geq \sum_{i \in I_+} w_i (a \sum_{x \in \bar{\mathbf{x}}_i} h(x) + b) - \sum_{i \in I_-} w_i (c \sum_{x \in \bar{\mathbf{x}}_i} h(x) + d) \qquad (33) \\
&= \sum_{i \in I_+} w_i a \sum_{x \in \bar{\mathbf{x}}_i} h(x) - \sum_{i \in I_-} w_i c \sum_{x \in \bar{\mathbf{x}}_i} h(x) + \sum_{i \in I_+} w_i b - \sum_{i \in I_-} w_i d. \qquad (34)
\end{aligned}$$

line (33) follows from Eq. (31). As evident by steps (1,2) of `MILearn`, In the sample $S_I$ all instances from positive bags have weight $\alpha(+1) = a$, and all instances from negative bags have weight $\alpha(-1) = c$. Therefore

$$\Gamma(h, S_I) = \sum_{i \in [m]} \sum_{x \in \bar{\mathbf{x}}_i} w_i y_i \alpha(y_i) h(x) = \sum_{i \in I_+} w_i a \sum_{x \in \bar{\mathbf{x}}_i} h(x) - \sum_{i \in I_-} w_i c \sum_{x \in \bar{\mathbf{x}}_i} h(x).$$

Combining this equality with Eq. (34) we get

$$\Gamma(\overline{h},\overline{S}) \geq \Gamma(h,S_I) + \sum_{i\in I_+} w_i b - \sum_{i\in I_-} w_i d.$$

Since $\sum_{i\in I_+} w_i = W_+$ and $\sum_{i\in I_-} w_i = W_- = 1 - W_+$, it follows that

$$\Gamma(\overline{h},\overline{S}) \geq \Gamma(h,S_I) + bW_+ - dW_- = \Gamma(h,S_I) + (b+d)W_+ - d. \tag{35}$$

Now, for any hypothesis $h$ we can conclude from Eq. (32) that

$$\begin{aligned}\Gamma(h,S_I) &= \sum_{i\in I_+} aw_i \sum_{x\in\bar{\mathbf{x}}_i} h(x) - \sum_{i\in I_-} cw_i \sum_{x\in\bar{\mathbf{x}}_i} h(x)\\ &\geq \sum_{i\in I_+} aw_i(\overline{h}(\bar{\mathbf{x}}_i) - d)/c - \sum_{i\in I_-} cw_i(\overline{h}(\bar{\mathbf{x}}_i) - b)/a\\ &= \sum_{i\in I_+} \frac{a}{c} w_i \overline{h}(\bar{\mathbf{x}}_i) - \sum_{i\in I_-} \frac{c}{a} w_i \overline{h}(\bar{\mathbf{x}}_i) - \sum_{i\in I_+} adw_i/c + \sum_{i\in I_-} cbw_i/a\\ &= \frac{c}{a}\Gamma(\overline{h},\overline{S}) + (\frac{a}{c} - \frac{c}{a}) \sum_{i\in I_+} w_i \overline{h}(\bar{\mathbf{x}}_i) - \frac{ad}{c}W_+ + \frac{cb}{a}W_-\\ &= \frac{c}{a}\Gamma(\overline{h},\overline{S}) + (\frac{a}{c} - \frac{c}{a}) \sum_{i\in I_+} w_i \overline{h}(\bar{\mathbf{x}}_i) - (\frac{ad}{c} + \frac{cb}{a})W_+ + \frac{cb}{a}.\end{aligned}$$

In the last equality we used the fact that $W_- = 1 - W_+$. Since $Z = \frac{c}{a}$, it follows that

$$\Gamma(h,S_I) \geq Z\Gamma(\overline{h},\overline{S}) + (\frac{1}{Z} - Z) \sum_{i\in I_+} w_i \overline{h}(\bar{\mathbf{x}}_i) - (\frac{d}{Z} + Zb)W_+ + Zb. \tag{36}$$

We will now lower-bound the right-hand-side of Eq. (36). Note that $\frac{1}{Z} - Z \leq 0$ since $c \geq a$. Therefore we need an upper bound for $\sum_{i\in I_+} w_i \overline{h}(\bar{\mathbf{x}}_i)$. We consider each of the two cases in the statement of the lemma separately.

**Case 1: $\mathcal{A}$ is $\epsilon$-optimal** We have $\sum_{i\in I_+} w_i \overline{h}(\bar{\mathbf{x}}_i) \leq \sum_{i\in I_+} w_i = W_+$. Therefore, by Eq. (36) for any $h \in \mathcal{H}$

$$\Gamma(h,S_I) \geq Z\Gamma(\overline{h},\overline{S}) + (\frac{1}{Z} - Z - \frac{d}{Z} - Zb)W_+ + Zb. \tag{37}$$

For a natural $n$, set $h_*^n$ such that $\Gamma(\overline{h}_*^n,\overline{S}) \geq \gamma^* - \frac{1}{n}$. We have (see explanations below)

$$\begin{aligned}\Gamma(\overline{h}_I,\overline{S}) &\geq \Gamma(h_I,S_I) + (b+d)W_+ - d &(38)\\ &\geq \Gamma(h_*^n,S_I) + (b+d)W_+ - d - \epsilon W &(39)\\ &\geq Z\Gamma(\overline{h}_*^n,\overline{S}) + (\frac{1}{Z} - Z - \frac{d}{Z} - Zb)W_+ + Zb + (b+d)W_+ - d - \epsilon W &(40)\\ &= Z\Gamma(\overline{h}_*^n,\overline{S}) + (\frac{1}{Z} - Z + (1-\frac{1}{Z})(d - Zb))W_+ + Zb - d - \epsilon W\\ &\geq Z(\gamma^* - \frac{1}{n}) + (\frac{1}{Z} - Z + (1-\frac{1}{Z})(d - Zb))W_+ + Zb - d - \epsilon W.\end{aligned}$$

Eq. (38) is a restatement of Eq. (35). Eq. (39) follows from the $\epsilon$-optimality of $\mathcal{A}$. Eq. (40) follows from Eq. (37). By taking $n \to \infty$, this inequality proves case (1) of the lemma.

**Case 2: $\mathcal{A}$ is one-sided-$\epsilon$-optimal** We have $\sum_{i \in I_+} w_i \overline{h}(\bar{\mathbf{x}}_i) \leq \sum_{i \in I_+} w_i = W_+$. Let $\overline{h} \in \Omega(\overline{\mathcal{H}}, \overline{S})$. Then for all $i \in I_-$, $\overline{h}(\bar{\mathbf{x}}_i) = -1$. Therefore

$$\begin{aligned}\Gamma(\overline{h}, \overline{S}) &= \sum_{i \in I_+} w_i \overline{h}(\bar{\mathbf{x}}_i) - \sum_{i \in I_-} w_i \overline{h}(\bar{\mathbf{x}}_i) \\ &= \sum_{i \in I_+} w_i \overline{h}(\bar{\mathbf{x}}_i) + \sum_{i \in I_-} w_i \\ &= \sum_{i \in I_+} w_i \overline{h}(\bar{\mathbf{x}}_i) + W_-.\end{aligned}$$

Therefore $\sum_{i \in I_+} w_i \overline{h}(\bar{\mathbf{x}}_i) = \Gamma(\overline{h}, \overline{S}) - W_- = \Gamma(\overline{h}, \overline{S}) + W_+ - 1$. Combining this with Eq. (36) we get

$$\begin{aligned}\Gamma(h, S_I) &\geq Z\Gamma(\overline{h}, \overline{S}) + (\frac{1}{Z} - Z) \sum_{i \in I_+} w_i \overline{h}(\bar{\mathbf{x}}_i) - (\frac{d}{Z} + Zb)W_+ + Zb \\ &= Z\Gamma(\overline{h}, \overline{S}) + (\frac{1}{Z} - Z)(\Gamma(\overline{h}, \overline{S}) + W_+ - 1) - (\frac{d}{Z} + Zb)W_+ + Zb. \\ &= \frac{1}{Z}\Gamma(\overline{h}, \overline{S}) + (\frac{1}{Z} - Z - \frac{d}{Z} - Zb)W_+ + Zb - \frac{1}{Z} + Z. \qquad (41)\end{aligned}$$

For a natural $n$, set $\overline{h}_+^n \in \Omega(\overline{\mathcal{H}}, \overline{S})$ such that $\Gamma(\overline{h}_+^n, \overline{S}) \geq \gamma_+^* - \frac{1}{n}$. For all bags $i \in I_-$, $\overline{h}_+^n(\bar{\mathbf{x}}_i) = -1$. Thus $\psi(h_+^n(x_i[1]), \ldots, h_+^n(x_i[|\bar{\mathbf{x}}_i|])) = -1$. By the assumption on $\psi$ in case (2) of the lemma, this implies that for all $i \in I_-, j \in [|\bar{\mathbf{x}}_i|]$, $h_+^n(x_i[j]) = -1$. Therefore $h_+^n \in \Omega(\mathcal{H}, S_I)$. We have (see explanations below)

$$\begin{aligned}\Gamma(\overline{h}_I, \overline{S}) &\geq \Gamma(h_I, S_I) + (b+d)W_+ - d \qquad (42) \\ &\geq \Gamma(h_+^n, S_I) + (b+d)W_+ - d - \epsilon W \qquad (43) \\ &\geq \frac{1}{Z}\Gamma(\overline{h}_+^n, \overline{S}) + (\frac{1}{Z} - Z - \frac{d}{Z} - Zb)W_+ + Zb - \frac{1}{Z} + Z + (b+d)W_+ - d - \epsilon W \qquad (44) \\ &= \frac{1}{Z}\Gamma(\overline{h}_+^n, \overline{S}) + (\frac{1}{Z} - Z + (1 - \frac{1}{Z})(d - Zb))W_+ + Zb - d + Z - \frac{1}{Z} - \epsilon W \\ &\geq \frac{1}{Z}(\gamma_+^* - \frac{1}{n}) + (\frac{1}{Z} - Z + (1 - \frac{1}{Z})(d - Zb))W_+ + Zb - d + Z - \frac{1}{Z} - \epsilon W.\end{aligned}$$

Eq. (42) is a restatement of Eq. (35). Eq. (43) follows from the one-sided-$\epsilon$-optimality of $\mathcal{A}$ and the fact that $h_+^n \in \Omega(\mathcal{H}, S_I)$. Eq. (44) follows from Eq. (41). By considering $n \to \infty$, this proves the second part of the lemma. ■

**Proof** [of Theorem 27] `MILearn` selects the hypothesis with the best edge on $\overline{S}$ between $\overline{h}_I$ and $\mathbf{h}_{\mathrm{pos}}$. Therefore

$$\Gamma(h_\circ, \overline{S}) = \max(\Gamma(\mathbf{h}_{\mathrm{pos}}, \overline{S}), \Gamma(\overline{h}_I, \overline{S})).$$

We have

$$\Gamma(\mathbf{h}_{\mathrm{pos}}, \overline{S}) = \sum_{i \in [m]} w_i y_i \mathbf{h}_{\mathrm{pos}}(\bar{\mathbf{x}}_i) = \sum_{i \in [m]} w_i y_i = W_+ - W_- = 2W_+ - 1.$$

Thus

$$\Gamma(h_\circ, \overline{S}) = \max(2W_+ - 1, \Gamma(\overline{h}_I, \overline{S})). \tag{45}$$

We now lower-bound $\Gamma(h_\circ, \overline{S})$ by bounding $\Gamma(\overline{h}_I, \overline{S})$ separately for the two cases of the theorem. Let $W$ be the total weight of $S_I$. Since $R \subseteq [r]$, $a \leq c$, and $\sum_{i\in[m]} w_i = 1$, we have

$$W = \sum_{i:y_i=+1} \sum_{x\in\bar{\mathbf{x}}_i} aw_i + \sum_{i:y_i=-1} \sum_{x\in\bar{\mathbf{x}}_i} cw_i \leq rc \sum_{i\in[m]} w_i = rc \tag{46}$$

**Case 1: $\mathcal{A}$ is $\epsilon$-optimal** From Lemma 32 and Eq. (46) we have

$$\begin{aligned}
\Gamma(\overline{h}_I, \overline{S}) &\geq Z\gamma^* + (\frac{1}{Z} - Z + (1 - \frac{1}{Z})(d - Zb))W_+ + Zb - d - rc\epsilon \\
&= Z\gamma^* + (\frac{1}{Z} - Z + (1 - \frac{1}{Z})(Z - 1 + \eta))W_+ - (Z - 1 + \eta) - rc\epsilon \\
&= Z\gamma^* + (\eta - 2)(1 - \frac{1}{Z})W_+ + 1 - \eta - Z - rc\epsilon.
\end{aligned}$$

The second line follows from the assumption $d - Zb - Z + 1 = \eta$. Combining this with Eq. (45) we get

$$\Gamma(h_\circ, \overline{S}) \geq \max\{2W_+ - 1,\ Z\gamma^* + (\eta - 2)(1 - \frac{1}{Z})W_+ + 1 - \eta - Z - rc\epsilon\}.$$

The right-hand-side is minimal when the two expressions in the maximum are equal. This occurs when

$$W_+ = W_\circ \triangleq \frac{Z\gamma^* + 2 - \eta - Z - rc\epsilon}{2 + (2 - \eta)(1 - \frac{1}{Z})}.$$

Therefore, for any value of $W_+$

$$\Gamma(h_\circ, \overline{S}) \geq 2W_\circ - 1 = \frac{Z\gamma^* - Z + \frac{1}{Z} - \frac{\eta}{2}(1 + \frac{1}{Z}) - rc\epsilon}{1 + (1 - \frac{\eta}{2})(1 - \frac{1}{Z})}.$$

**Case 2: $\mathcal{A}$ is one-sided-$\epsilon$-optimal** From Lemma 32 and Eq. (46) we have

$$\begin{aligned}
\Gamma(\overline{h}_I, \overline{S}) &\geq \frac{1}{Z}\gamma_+^* + (\frac{1}{Z} - Z + (1 - \frac{1}{Z})(d - Zb))W_+ + Zb - d + Z - \frac{1}{Z} - rc\epsilon \\
&= \frac{1}{Z}\gamma_+^* + (\frac{1}{Z} - Z + (1 - \frac{1}{Z})(Z - 1 + \eta))W_+ - (Z - 1 + \eta) + Z - \frac{1}{Z} - rc\epsilon \\
&= \frac{1}{Z}\gamma_+^* + (\eta - 2)(1 - \frac{1}{Z})W_+ + 1 - \eta - \frac{1}{Z} - rc\epsilon.
\end{aligned}$$

The second line follows from the assumption $d - Zb = Z - 1 + \eta$. Combining this with Eq. (45) we get

$$\Gamma(h_\circ, \overline{S}) \geq \max\{2W_+ - 1,\ \frac{1}{Z}\gamma_+^* + (\eta - 2)(1 - \frac{1}{Z})W_+ + 1 - \eta - \frac{1}{Z} - rc\epsilon\}.$$

The right-hand-side is minimal when the two expressions in the maximum are equal. This occurs when

$$W_+ = W_\circ \triangleq \frac{\gamma_+^* - 1 + (2 - \eta - rc\epsilon)Z}{2Z + (2 - \eta)(Z - 1)}.$$

Substituting $W_+$ for $W_\circ$ in the lower bound, we get

$$\Gamma(h_\circ, \overline{S}) \geq 2W_\circ - 1 = \frac{\gamma_+^* - \frac{\eta}{2}(Z+1) - rc\epsilon Z}{2Z - 1 - \frac{\eta}{2}(Z-1)}.$$

■